\documentclass[lettersize,journal]{IEEEtran}

\usepackage{booktabs}
\usepackage{diagbox}
\usepackage{tablefootnote}
\usepackage{footnote}
\usepackage{multirow}
\usepackage{algorithm}
\usepackage{algpseudocode}
\usepackage{amsmath}
\usepackage{graphics}
\usepackage[utf8]{inputenc}
\usepackage{epsfig}

\newcommand{\tabincell}[2]{\begin{tabular}{@{}#1@{}}#2\end{tabular}}

\newtheorem{Assm}{Assumption}

\usepackage{subfigure}
\usepackage{helvet}  
\usepackage{courier}  
\usepackage{url}  
\usepackage{graphicx}  
\usepackage{makecell}
\usepackage{footnote}
\usepackage{tablefootnote}
\usepackage{enumerate}

\usepackage{algpseudocode}
\usepackage{amsmath,amssymb,amsfonts}

\usepackage{algorithmicx}
\usepackage{algorithm}
\usepackage{algpseudocode}
\usepackage{bm}
\usepackage{marvosym}
\usepackage{graphics}

\usepackage{url}
\usepackage{color}
\usepackage{epstopdf}
\usepackage{subfigure}

\usepackage{breqn}
\usepackage{multirow}
\usepackage{booktabs}
\usepackage{subfigmat}
\usepackage{subfig}

\usepackage{url}

\begin{document}
	
	\title{ComS2T: A complementary spatiotemporal learning system for data-adaptive model evolution}
	
	\author{Zhengyang~Zhou$^\dagger$,~\IEEEmembership{Member,~IEEE,}
		Qihe Huang$^\dagger$, 
		Binwu~Wang,
		Jianpeng Hou,
		Kuo~Yang,\\
		Yuxuan~Liang,\IEEEmembership{ Member,~IEEE}, ~Yu Zheng,\IEEEmembership{ Fellow,~IEEE,}
		Yang~Wang$^*$, \IEEEmembership{Senior Member,~IEEE.}
		\IEEEcompsocitemizethanks{\IEEEcompsocthanksitem Z. Zhou,  Q. Huang, B. Wang, J. Hou, K. Yang, and Y. Wang$^*$ are with University of Science and Technology of China, China.  Z. Zhou and Y. Wang are also with Suzhou Institute for Advanced Research, USTC, Suzhou, China.  Y. Liang is with Hong Kong University of Science and Technology, Guangzhou, China.  Y. Zheng is with JD.COM and JD Intelligent Cities Research, Beijing, China. \protect\\
			E-mail: {\{ zzy0929, angyan$^*$\}@ustc.edu.cn,\{hqh, wbw1995, Enterpr1se, yangkuo\}@mail.ustc.edu.cn,  yuxliang@outlook.com, msyuzheng@outlook.com.}
			\IEEEcompsocthanksitem{$^\dagger$Equal contribution.}  
			\IEEEcompsocthanksitem{$^*$Prof. Yang Wang is the corresponding author.}	
		}
		
		
		\thanks{Manuscript received Feb 25th, 2024; revised xx xxth, 2024.}}
	
	\markboth{Journal of \LaTeX\ Class Files,~Vol.~14, No.~8, August~2021}%
	{Shell \MakeLowercase{\textit{et al.}}: A Sample Article Using IEEEtran.cls for IEEE Journals}
	
	
	\maketitle
	
	\begin{abstract}
	Spatiotemporal (ST) learning has become a crucial technique to enable smart cities and sustainable urban development. Current ST learning models capture the heterogeneity via various spatial convolution and temporal evolution blocks. However, rapid urbanization leads to fluctuating distributions in urban data and city structures over short periods, resulting in existing methods suffering generalization and data adaptation issues. Despite efforts, existing methods fail to deal with newly arrived observations and those methods with generalization capacity are limited in repeated training. Motivated by complementary learning in neuroscience, we introduce a prompt-based complementary spatiotemporal learning termed ComS2T, to empower the evolution of models for data adaptation. ComS2T partitions the neural architecture into a stable neocortex for consolidating historical memory and a dynamic hippocampus for new knowledge update. We first disentangle two disjoint structures into stable and dynamic weights, and then train spatial and temporal prompts by characterizing distribution of main observations to enable prompts adaptive to new data. This data-adaptive prompt mechanism, combined with a two-stage training process, facilitates fine-tuning of the neural architecture conditioned on prompts, thereby enabling efficient adaptation during testing. Extensive experiments validate the efficacy of ComS2T in adapting to various spatiotemporal out-of-distribution scenarios while maintaining efficient inference capabilities.
		
	\end{abstract}
	
	\setcounter{secnumdepth}{6}
	\section{Introduction}
\begin{figure}[ht]	
		\centering
		\includegraphics[width=\linewidth]{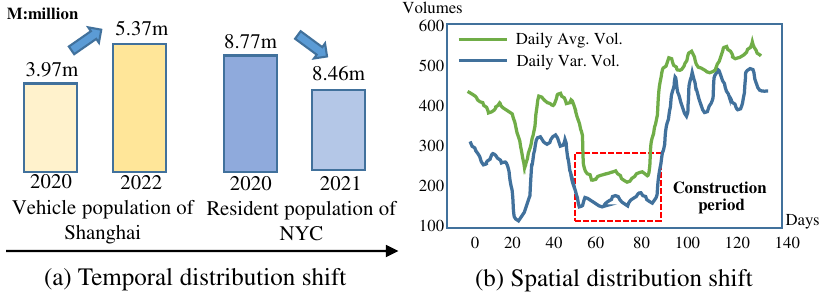}
		\caption{Examples of urban dynamics along city evolution}
		\label{fig:urbandynamics}
	\end{figure}

	\IEEEPARstart{S} patiotemporal (ST) learning, which empowers intelligent management decisions, has become a pivotal technique to improve the quality of urban life as well as the intelligence of cities. Current spatiotemporal forecasting models usually incorporate various spatial convolution blocks and temporal dependence extractors to achieve predictions, enabling diverse multi-variate urban series forecasting ~\cite{wu2020connecting,huang2023crossgnn,zheng2023diffuflow,jin2023spatio,wang2020deep,lu2022spatio,pan2020spatio,liu2023itransformer,dong2024simmtm,wu2022timesnet,liu2022non}, including traffic conditions~\cite{zhou2020foresee,ji2023spatio,zheng2023diffuflow,li2018diffusion,zhou2020riskoracle}, natural climates~\cite{wu2023interpretable,castro2021stconvs2s,chen2023fengwu,zhang2023skilful}, as well as  key indexes of environments~\cite{liang2023airformer,du2023deciphering,du2019deep,chen2023group}. Despite prosperity, most existing methods assume that training and testing data are both independent and identically distributed (i.i.d.) where the principle does not hold on in real-world scenarios. Actually, urban spatiotemporal elements tend to expand and increase with the urbanization and city evolution. In Fig.~\ref{fig:urbandynamics}, we take human mobility prediction as an instance. From a macro view, the vehicle population of Shanghai increases to 5.37 million in 2022 from 3.97 million in 2020, while the demographic population of NYC decrease from 8.77 million to 8.46 million, from 2020 to 2021. In a micro view, if one region experiences the construction of a shopping mall, the mobility intensity will go through a sudden decrease during the construction period, following an increase after constructed. 
	As a result, the  shifts regarding temporal distribution  and urban  structures   pose  out-of-distribution (OOD) challenges on respective temporal  and spatial perspectives to current ST models. Therefore,  a data-adaptive spatiotemporal learning framework with timely model update  becomes  highly required.

	Although majority of spatiotemporal learning methods fail to adapt their models to new OOD instances,  learning graphs with OOD settings have increasingly raised the attention of researchers~\cite{wang2023brave,wu2021discovering,wu2021handling,li2022let}. In general, environment is a fundamental concept in OOD learning where researchers can explicitly capture the invariance across environments for transfer. In our spatiotemporal learning scheme, we can also inherit the environment concept and classify spatiotemporal environments into temporal aspect and local urban structures. To address temporal distribution shifts, AdaRNN first defines covariate shifts in time series and designs distribution characterization to aggregate previous series for weighted prediction~\cite{du2021adarnn}. Following AdaRNN, CauSTG reflects complex spatial-temporal dependencies by learnable parameters and takes invariant learning in causal perspective as a priori~\cite{zhou2023maintain}, then transfer stable weights from historical environment to unseen scenarios. Unfortunately, CauSTG only captures stable neural structure but fail to improve the dynamic structure for data adaptation. On emerging urban structures, pioneering works take continuous learning into consideration. TrafficStream updates the neural structure by identifying the most dissimilar new nodes and consolidate historical information with experience reply~\cite{chen2021trafficstream}, while PECPM manages the ST pattern bank with newly involved nodes, reducing the memory burdens~\cite{wang2023pattern}. Very recently, sUrban integrates contrastive learning with invariant learning to disentangle the environment-invariant and environment-aware representation for OOD transfer~\cite{wang2023surban}. However, experience-reply or memory-based solutions require unlimited data space and multiple model re-training, leading to increased computational and storage spaces. Modeling data distributions from environments~\cite{xia2023deciphering,wang2023surban, yuan2023environment} can actually empower generalization by imitating the perturbation on features and extending the boundary of training set thus enlarging the receptive field of models. Even though, they still fall into learning with closed training set, which has difficulty in dealing with new instances. 
	
	Therefore, we can summarize two serious problems in existing solutions to OOD challenges. First, these methods explicitly model environments but still have nothing to do with the new arrived data, especially lack the designs to accommodate  model evolution and data-adaptive update~\cite{zhou2023maintain,xia2023deciphering, yuan2023environment}. Second, current frameworks including continuous learning, suffer computational burdens and space complexity to preserve the historical regularity~\cite{zhou2023maintain} and  new patterns~\cite{chen2021trafficstream,wang2023pattern}, thus limiting the efficiency of model generalization when the patterns and structures vary.
	
	Fortunately, there is some update on the cognition of memory mechanism in human brain, i.e., different regions in our brain usually carry out distinctive roles and work in a complementary manner to consolidate  historical memory and assimilate fresh knowledge~\cite{o2014complementary,kumaran2016learning,mcclelland2020integration,lee2020clinical}. In particular, it reveals that the neocortex neural module gradually acquires structured and well-learned historical knowledge while the hippocampus structure tends to efficiently learn specific individual instance-level skills~\cite{arani2022learning,kumaran2016learning}. Such insight, formally designated as complementary learning system (CLS), opportunely provides clues to consolidating and updating model parameters in a complementary perspective to  adapt streaming and dynamic spatiotemporal observations. 
	
	Recently, the complementary learning system  has already been investigated to realize continuous learning~\cite{arani2022learning}, but it still has never been coupled with spatiotemporal frameworks. Given that spatiotemporal observations reveal complex spatial and temporal dependencies with interactions between environmental factors, while CLS requires disjoint learning structures and effective update strategy, thereby the following challenges remain unresolved. 
	\begin{itemize}
		\item How to seamlessly couple the complex spatiotemporal learners with complementary learning in a unified and efficient framework, i.e., given a ST learner, how to efficiently identify both stable neocortex neural module  and dynamic hippocampus structure for transferability and model update, respectively?
		\item How to cooperatively model spatiotemporal observations with  environment features in a holistic perspective, and then appropriately deal with unseen data to adapt the hippocampus structure to new environments? 
		\item How to design the training strategy to simultaneously preserve the historical information and endow online model update upon new patterns with limited consumption? 
	\end{itemize}
	In this work, inspired from complementary learning system in neuroscience, we propose a prompt-based~\cite{zhang2023promptst,liu2022p} data adaptive Complementary ST learning System (ComS2T) to tackle the OOD challenge and endow the model evolution capacity. Our ComS2T actively identifies the respective stable and dynamic subspace of learning weights to instantiate the complementary learning. Firstly, by reflecting the spatial-temporal dependencies into learnable parameters, we disentangle the full learnable neural weights into two complementary subspaces, stable neocortex and dynamic hippocampus by preserving two variation matrices for capturing the dynamics of weight behaviors along the learning process. Secondly, to disentangle and refine the environment-observation interactions, we take spatial location description and temporal signals as basic environment signals for following prompts, and  train learnable spatial-temporal prompts by characterizing the input main observations with parameterized distribution. We then exploit the well-learned prompts to fine-tune hippocampus structures of our spatial-temporal blocks, allowing the whole architecture to evolve with new input observations. Finally, we devise a two-stage training process with spatiotemporal warm-up and prompt-based fine-tune, which progressive learns the mapping functions conditional on prompts and allows efficient adaptation during testing stage. To this end, our complementary learning empowers model evolution on both training and testing stages. Specifically, along the training process, our CLS simultaneously  preserves the historical information and allows the flexibility of hippocampus. During  testing process, the designs of adapting spatial-temporal prompts  to testing data with limited self-training further enables the model adaption. The contributions can be achieved as follows.
	\begin{itemize}
		\item It is the first attempt to couple complementary learning in neuroscience with spatiotemporal models to realize  generalization and data adaptation, where an efficient neural architecture disentanglement is devised through two well-preserved variation matrices.
		\item A self-supervised prompt training is proposed to bridge the gap between environment factors and distribution of main observation, which  not only allows prompts for neural network fine-tune, but also enables the dynamics and evolution of model parameters sensitive to data distributions.
		\item Our framework can simultaneously deal with  shifts on both spatial and temporal aspects, and four OOD scenarios are constructed to imitate the data adaptation for model verification. Experiments show that our ComS2T can improve performances from 0.73\% to 20.70\% under temporal shifts, while promote 0.36\% to 17.30\% under structural shifts. 
	\end{itemize}
	
	\section{Preliminaries}
	\subsection{Problem formulation and basic structures}
	Given a sequence of spatiotemporal graphs with $T$ steps, $\mathbb{G} = \{\mathcal{G}_1, \mathcal{G}_2,...,  \mathcal{G}_t, ..., \mathcal{G}_T\}$,  each $\mathcal{G}_t$ is described as $\{\mathcal{V}, \bm{X}_t, \mathcal{E} \}$ where $\mathcal{V}=\{v_1, v_2,..., v_N\}$ is the node set, and  $\mathcal{E}$ describes the graph structure.  In the observed spatiotemporal graph, we denote  $\bm{X}_t \in \mathbb{R}^{N \times F}$ as the deterministic main observations of $\mathcal{G}_t$, while take ${\bf{E}}=\{ {\bf{E}}_s, {\bf{E}}_t \}$ as the observed contextual environments, consisting of spatial environment ${\bm{e}}_s \in {\bf{E}}_s$ such as geographical encoding and location index,  and temporal environment ${\bm{e}}_t \in {\bf{E}}_t$, such as day of week, timestamps of day, etc. Spatiotemporal learning aims to predict next consecutive $l$ steps by exploiting previous $\kappa$ steps, i.e., $\widehat{\bm{Y}} =f(\bm{X})$ where $(\bm{X},\bm{Y}) = ({\bm{X}_{t - \kappa+1 :t}},{\bm{X}_{t + 1:t+ l }})$.
	Generally, for a spatiotemporal learning framework $ f(\cdot)$, it usually consists of two main components, graph-based spatial learning ${F_S}(\bm{X})$ and temporal convolution module ${\Gamma _T(\bm{X}_S)}$, where the two components can learn alternately. Given input sequential observations $\mathbb{X} = \{ {\bm{X}_1},{\bm{X}_2},...,{\bm{X}_\kappa}\} $, we can formalize the spatial representation ${\bm{X}_s}$, the  output of spatial learning block $F_s(\cdot)$ as, 
	\begin{equation}
		{\bm{X}_S}{\text{ = }}{F_S}(\bm{X}){\text{ = GCN(}}\bm{A}{\bm{X} }{\bm{W}_{sp}}{{)}}
	\end{equation} 
	The output of temporal learning block is ${\bm{X}_{ST}}$ which is constructed by feeding $\bm{X}_S$ into ${\Gamma _T}$, i.e.,
	\begin{equation}
		{\bm{X}_{ST}} ={\Gamma _T} (\bm{X}_{S})= \text{TCN}({\bm{X}_S};{\bm{W}_T})
	\end{equation}
	where ${\bm{W}_s} = \{\bm{A},{\bm{W}_{sp}}\} $ and ${\bm{W}_T}$ respectively account for learning sets on spatial  and temporal perspectives.
	Given training and testing data $\mathcal{D}_{\mathit{train}} $, $\mathcal{D}_{\mathit{test}}$, 
	the data distribution shift refers to the changes of distribution over training and testing observations, i.e., ${P_{train}}(\bm{X}) \ne {P_{test}}(\bm{X})$, and the goal for data-adaptive 
	model evolution is to derive a new mapping $f^*$ simultaneously containing an invariant relation component and a data-adaptive dynamic component. 
	\begin{equation}
		\min \mathop \mathcal{R}\limits_{(\bm{x},\bm{y}) \in {\mathbb{G}_s}} (\widehat{\bm{y}};{f^*}(\bm{x}))
	\end{equation}
	
	\subsection{Theoretical analysis for complementary spatiotemporal learning}
	\label{sec:theogan}
	We first make the fundamental assumptions to our work, and then respectively illustrate the failure of traditional spatiotemporal learning on OOD regression, and deliver the superiority of our  complementary learning scheme with both invariance and dynamics. We will follow the assumptions in EERM~\cite{wu2021handling} and CauSTG~\cite{zhou2023maintain} to  facilitate our analysis. 
	
	\subsubsection{Fundamental assumptions}
	{\em \begin{Assm}[\textbf{Dependence between main observation and their environments}] Given sequential spatiotemporal observations $\bm{X}_1,\bm{X}_2, ..., \bm{X}_N$, we suppose the distributions ${\bf{P}}(\bm{X})$ can be dependent on the contextual environments ${\bf{E}}=\{ {\bf{E}}_s, {\bf{E}}_t \}$. \end{Assm}}
	
	Different from modeling categorical environments in other literature~\cite{xia2023deciphering,yuan2023environment}, we do not assume a closed set for environments but they are instantiated with continuous geographic and timestamp embeddings, indicating the urban structure and overall temporal shifts. Then the distribution shifts over   $\bm{X}_i$ can be attributed to changes of virtual environment  ${\bf{E}}$.

	{\em \begin{Assm}[\textbf{Invariance property}] Even though the  covariate distributions are changing over  environments, there must exist some invariant relations. Given two environments $\bm{e}_i, \bm{e}_j \in {\bf{E}}$, $ \exists (p,q), s.t.\; {\bf{P}}({x_p},{x_q}|\bm{e}_i) = {\bf{P}}({x_p},{x_q}|\bm{e}_j)$, where $x_p$ and $x_q$ are specific observations of $\bm{X}$. To this end, we can decompose all the relations between input $\bm{X}$ and $\bm{Y}$ into invariant parts and dynamic  parts,  accounting for respective causal and non-causal components.
	\end{Assm}}

	In the dynamic graph regression, given node $v_i$, let degree of $v_i$, and proportions of neighbors with causally invarint relations to $v_i$ denote as $d_i, p_i$ with $d_i>1, 0<p_i<1$. Considering the covariate shifts, i.e., we train the model on samples following Gaussian distribution ${\mathbb{G}_s} \sim {N}({\mu _0},{\sigma _0}|{e_0})$ and test on samples following  ${\mathbb{G}_t} \sim {N}({\mu _q},{\sigma _q}|{e_q})$.

	\begin{figure*}[ht]	
		\centering
		\includegraphics[width=\linewidth]{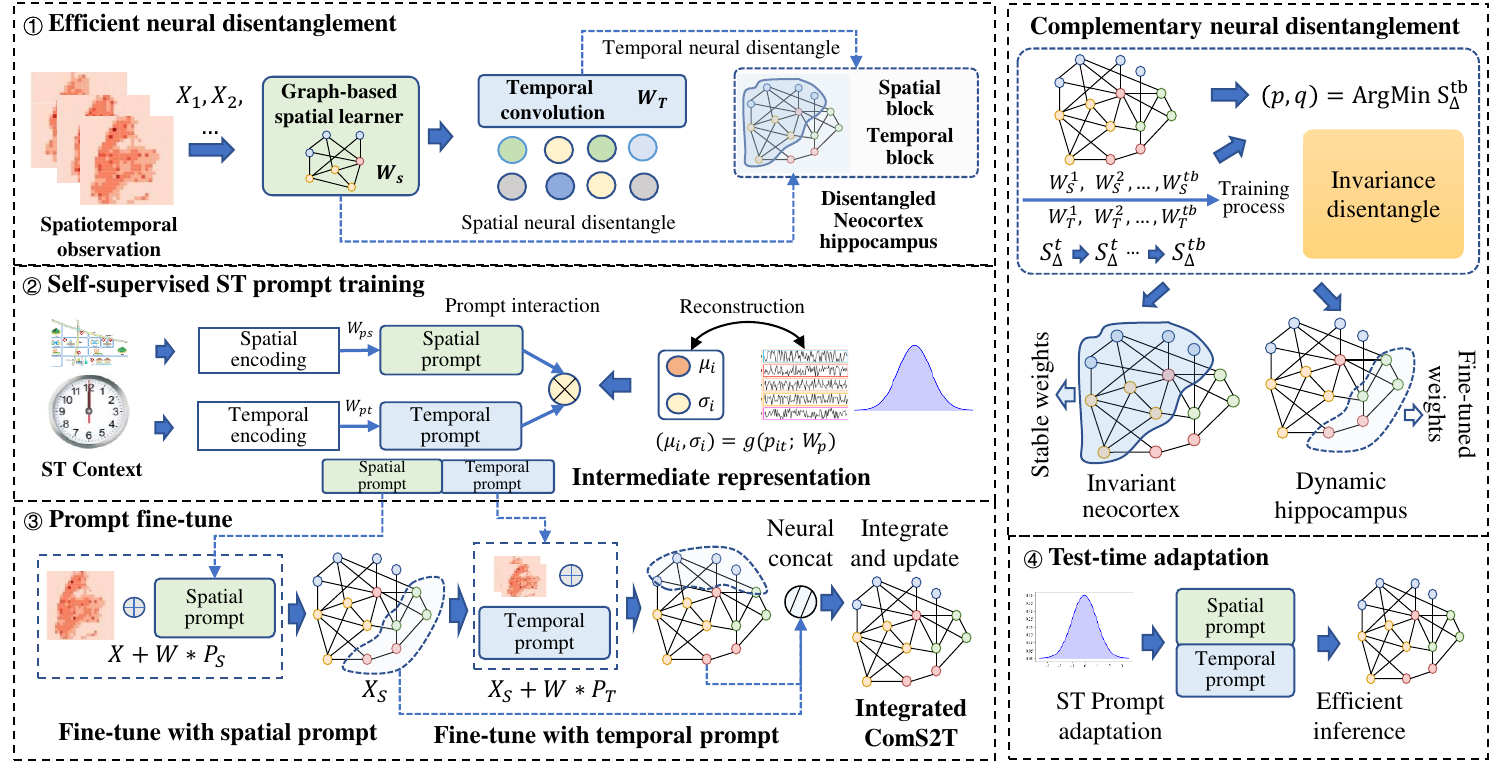}
		\caption{Framework overview of ComS2T, consisting of four major components.  The right top {Complementary neural disentanglement} is the detailed illustration of the first component for decoupling stable and dynamic weight spaces in CLS. }
		\label{fig:FO}
	\end{figure*}
	\subsubsection{Failure on traditional learning}
	For GNN-based representation learning, we consider one-time aggregation from $T$-step to achieve the expected regression prediction of ${T+1}$-step. Based on above assumptions, for node $v_i$, we take $\mathcal{N}_c(v_i)$ as the causally correlated neighbor set of $v_i$ while  $\mathcal{N}_s(v_i)$ denotes the set of non-causally correlated neighbors. Given degree $d_i$, the traditional one-time aggregation for $v_i$ with all nodes can be formulated by decomposing the causal and non-causal parts, i.e.,
	\begin{equation}
		E(h_i^T) = \frac{{x_i^T + \sum\limits_{{c_j} \in {\mathcal{N}_c}({v_i})} {{w_{ij}^{c}}x_{{c_j}}^T}  + \sum\limits_{{s_j} \in {\mathcal{N}_s}({v_i})} {{w_{ij}^{s}}x_{{s_j}}^T} }}
		{{1 + {d_i}}}
		\label{eq:aggr}
	\end{equation}
	where $c_j$ and $s_j$ are the subscripts of two neighborhood sets, and  ${w_{ij}^{c}}$ and ${w_{ij}^{s}} $ are learnable weights for causal parts and non-causal parts.
	By calculating the difference between the aggregated expectation $E(h_i^T)$ and groundtruth $x_i^{T + 1}$, we can derive the prediction error ${\varepsilon _0}$ after one-time aggregation by neglecting the non-linear activations, i.e.,
	\begin{equation}
		\small
		\begin{split}
			{\varepsilon _0} & = ||E(h_i^T) - x_i^{T + 1}||\\
			& = ||\frac{{x_i^T + \sum\limits_{{c_j} \in {\mathcal{N}_c}({v_i})} {{w_{ij}^{c}}x_{{c_j}}^T}  + \sum\limits_{{s_j} \in {\mathcal{N}_s}({v_i})} {{w_{ij}^{s}}x_{{s_j}}^T}  - (1 + {d_i})x_i^{T + 1}}}
			{{1 + {d_i}}}  ||
		\end{split}
		\label{eq:error01}
	\end{equation}
	
	Assume that observations on both current step $x_i^T$ and next step $x_i^{T + 1}$  follow the same Gaussian distribution $N({\mu _0},{\sigma _0})$, and ${p_i} = \frac{{||{\mathcal{N}_c}({v_i})||}} {{||\mathcal{N}({v_i})||}}$ accounts for the proportion of causal neighbors. To facilitate the expression, we let $\mu _0^t,\mu _0^{t + 1}$ denote the expectation of observation $x_i$ at $t$ and $t+1$, and  $\mu _0^c,\mu _0^s$
	represent the expectation of the expected observation  of its causal neighborhood and non-causal (spurious) neighborhood. The initial error of Eq.(\ref{eq:error01}) can be modified into,
	\begin{equation}
		{\varepsilon _0} = \frac{{\mu _0^t + {p_i}{d_i}\mu _0^cw_i^c + (1 - {p_i}){d_i}\mu _0^sw_i^s - (1 + {d_i})\mu _0^{t + 1}}}
		{{1 + {d_i}}}
		\label{eq:error02}
	\end{equation}
	where we ignore the sign for absolute value, and assume that the expectation and learnable weights all preserve positive.
	
	Since the non-causal based learning is formulated by regression function of ${{\widehat y}_i} = w_i^c{x_c} + w_i^s{x_s}$, the prediction residual ${\mathit{res}}_i$ can be derived by  ${\mathit{res}}_i = {{\widehat y}_i} - w_i^c{x_c} = w_i^s{x_s}$. Therefore, we can substitute the difference between aggregated causal parts and groundtruth with aggregated non-causal part, and obtain the following equation,
	\begin{equation}
		\begin{split}
			{\varepsilon _0} = \frac{{2(1 - {p_i}){d_i}\mu _0^sw_i^s}}
			{{1 + {d_i}}} \hfill \\ 
		\end{split} 
		\label{eq:error03}
	\end{equation}
	
	With Eq. (\ref{eq:error03}), we can arrive that the derived error is not reducible as  $w_i^s \ne 0$. The learnable weights in traditional spatiotemporal learning or regession problems are not separately considered, we thus we assume the distributions of corresponding learnable weights by, i.e., ${w_i^c, w_i^s \sim N({\mu _w},{\sigma _{w}})}$. 
	
	For the OOD testing set $N(\mu_q, \sigma_q|e_q)$ satisfying $\mu_q = q\mu_0 (q \in \mathbb{N}^+)$, the approximated error of OOD testing is amplified to,  
	\begin{equation}
		{\varepsilon _q}\sim \frac{{2(1 - {p_i}){d_i}q{\mu _0}{\mu _w}}}{{1 + {d_i}}}
	\end{equation}
	where $\mu _0^s$ is sampled from ${N}({\mu _0},{\sigma _0})$. 
	
	Therefore, with $\mu_w \ne 0$, this derivation manifests that learning over OOD scenarios suffer $q-times$ errors of in-distribution  (ID) ones, resulting in an unacceptable amplification of error bounds from ID to OOD samples.

	\subsubsection{Superiority of complementary learning}
	As CLS is a pioneering research line in machine learning, there is very few systematically theoretical analysis on it. We here provide a brief dissection along above GNN-based prediction.
	Given ${{\widehat {\bm{y}}}_i} = \bm{w}_i^c{\bm{x}_c} + \bm{w}_i^s{\bm{x}_s}$, our ComS2T first disentangles the invariant neural architecture and dynamic context-sensitive architectures, then the learnable weights can be explicitly separated into two sections, i.e., $\bm{W} = \{\bm{w}^c, \bm{w}^s\}$, for 
	stable neocortex and dynamic hippocampus structures. The two learnable parts are explicitly considered as satisfying the following distributions, i.e., 
	\begin{equation}
		\label{eq:sep_weigh}
		{\bm{w}^c\sim N({\mu _w},{\sigma _{wc}}),\;\;\bm{w}^s \sim N({\mu _w},{\sigma _{ws}})}
	\end{equation}
	In OOD scenarios of our CLS, $\bm{w}^c$ should be static thus transferable  across environments, while $\bm{w}^s$ can be variable and timely updated upon the distribution of main observation changes. Then given node $v_i$ in spatiotemporal graph, Eq.(\ref{eq:error02}) becomes reducible and can be suppressed by optimizing  $w_i^s$. Let ${\rm Eq}.(\ref{eq:error02}) = 0$, the  $w_i^s$ can have an analytical  solution to this optimization, shown as,
	\begin{equation}
		w_i^s = \frac{{(1 + {d_i})\mu _0^{t + 1} - (\mu _0^t + {p_i}{d_i}\mu _0^cw_i^c)}}
		{{(1 - {p_i}){d_i}\mu _0^s}}
	\end{equation}
	
	To this end, we conclude that our ComS2T can potentially converge to optimal results under the distribution shifts with a disentangled and update mechanism for data adaptation.

	\section{Methodology}
	\subsection{Framework overview}
	By taking insights of neuroscience, our ComS2T unifies the invariance and dynamics into a complementary spatiotemporal learning system. The complementary system can be interpreted as two aspects, it first efficiently disentangles the learnable neural weights into two complementary subspaces, stable neocortex and dynamic hippocampus by explicitly modeling learning behavior, where two structures work cooperatively to dynamically adapt streaming spatiotemporal data. Second, ComS2T  pre-trains the spatial-temporal prompts via self-supervised learning, bridging the gap between learnable prompts and specific data patterns and such pre-training strategy allows test-time training and model sensitivity to distribution shifts. Finally, our ComS2T devises a progressive learning architecture, consisting of four major components, efficient neuron disentanglement, prompt pre-training, prompt-based fine-tune and test-time self-adaptation, to realize the data adaptation from coarse-grained to fine-grained. 
	
	\subsection{Efficient neural disentanglement}
	\label{sec:invardis}
	The complementary learning in neuroscience reveals that  regions in  brain usually play different roles for remembering historical and fresh knowledge, where neocortex  acquires structured and well-learned historical knowledge while the hippocampus  tends to quickly learn specific  instance-level skills. In this section, we propose an efficient disentanglement to decouple the potential neocortex and hippocampus  structures from integrated neural network. 
	
	To ensure an interpretable and aligned disentanglement, we separately take spatial module and temporal module as separated units. 
	Assuming there are $K$ layers for spatial aggregation and $L$ times of  temporal convolution, we take both the spatial adjacency and feature-level scaling as the spatial learnable space ${\bm{W}_S} = {\{ {\bm{A}_i},{\bm{\omega}_{{s_i}}}\} _{i = 0,1,...,K}}$ and the set of temporal learnable weights  is designated as ${\bm{W}_T} = {\{ {\bm{w}_{{t_i}}}\} _{i = 0,1,...,L}}$.
	For easier notation, we  take all learnable parameters in $f(\cdot)$ as $ \bm{W} = \{ {\bm{W}_s},{\bm{W}_T}\}$, and denote $w_{s_{ij}}$ and $w_{t_{ij}}$ as the specific deterministic element in  ${\bm{W}_s} \in \mathbb{R}^{P\times Q}$ and  ${\bm{W}_T} \in \mathbb{R}^{M \times N}$, where $P\times Q$ and ${M \times N}$ are the virtual dimensions for these two weight sets, respectively.

	Existing OOD generalization solutions often take multiple models for ensembling~\cite{zhou2023maintain}, which takes substantial computational burden and memory space. 
	Actually, we argue that the  model behavior can be described by $\bm{W}^0, \bm{W}^1, \cdots,  \bm{W}^{tb}$, where $\mathit{tb}$ is the training unit consisting of several batches  or epochs in the learning process (e.g., we can take one epoch as a training process unit). The variations over different neuron-level elements can be observed when the learning process achieves relatively stability. To capture the evolution behavior along  training process, we propose a differential accumulation strategy. Specifically, we instantiate two matrices the absolute differential  values  $\Delta {\widetilde{\bm{W}}}^{tb}$, characterizing differences between  adjacent training unit,  and the accumulated differences $\bm{S}_{\Delta}^{tb}$, the summation of all variations. Given  the  unit number within training process $tb$, the  element-level model evolution can be iteratively characterized by,
	\begin{equation}
		\left\{ \begin{matrix} 
			\Delta {\widetilde{{\bm{W}}}}^{tb} = ||{\bm{W}^{tb}} - {\bm{W}^{tb - 1}}|| \hfill \cr 
			\bm{S}_{_\Delta }^{tb} = \Delta \widetilde{\bm{W}}^{tb} + \bm{S}_{_\Delta }^{tb - 1} \hfill \cr 
		\end{matrix}  \right.
		\label{eq:DeltaM}
	\end{equation}
	Note that $\bm{S}_{_\Delta }^0 = \bm{0}$ is the initialized status, and the absolute differential values on learnable parameters can directly outline the quantified variations on streaming training samples and alleviate the influence of signed  variation on final accumulation.   
	We  calculate Eq.(\ref{eq:DeltaM}) in an element-wise manner,  and impose it on both spatial learning space ${\bm{W}_s}$ and temporal convolution space ${\bm{W}_T}$ to  refine stable relations  with well alignment and interpretability. 
	As the smaller values within $S_{_\Delta }^t $ indicate the more stable relations along the training process while the largest ones represent the most dynamics, we separate them by taking the minimal-$\tau\%$ variations as a threshold  respectively over  the spatial and temporal blocks of the architecture. The $\tau$ is a hyperparamter, which indicates the model fluctuation over training samples and can be optimized empirically.
	To this end, the indexes for largest variations in weight space are considered as the neocortex neural module for capturing invariant and stable spatial-temporal correlations,    
	while the complementary set (indexes for remaining $(1-\tau\%)$ largest values within $\bm{S}_{_\Delta }^t $) of the neural network consists of the  hippocampus neural structure for data-adaptive update. Given the training unit $tb$, we  denote  $ {\bm{S}_{sp\Delta}^{tb}},{\bm{S}_{tp\Delta}^{tb}}$ as the summarized variations on spatial and temporal learning blocks, and then the  neocortex and hippocampus structures can be determined by highlighting the most stable learnable weights by,
	\begin{equation}
		\left\{ \begin{matrix} 
			{\{(p,q)\}_S} = \mathop {\arg \min }\limits_{\scriptstyle {\;\;\;\;\text{Min}} - \tau \%  \hfill \atop 
				\scriptstyle 1 \leqslant p \leqslant P,1 \leqslant q \leqslant Q \hfill}  (\{  {\bm{S}_{sp\Delta}^{tb}}(p,q)\}) \hfill \cr 
			{\{(m,n)\}_T} = \mathop {\arg \min }\limits_{\scriptstyle {\;\;\;\;\text{Min}} - \tau \%  \hfill \atop 
				\scriptstyle 1 \leqslant m \leqslant M,1 \leqslant n \leqslant N \hfill}  (\{  {\bm{S}_{tp\Delta}^{tb}}(m,n)\} )) \hfill \cr 
		\end{matrix}  \right.
		\label{eq:disetgle}
	\end{equation}
	The  index sets $\{(p,q)\}_S, \{(m,n)\}_T$ are index sets of selected neuron-level  neocortex elements in spatial  and temporal aspects  ${\bm{W}_s}$ and ${\bm{W}_T}$. After that, the disentanglement process can be implemented by, 
	\begin{equation}
		\left\{ \begin{matrix} 
			\bm{W}_{ne}^S = \mathop {{\text{Avg}}}\limits_{\scriptstyle {\;\;\text{    Min}} - \tau \%  \hfill \atop 
				\scriptstyle i,j \in \{ {(p,q)_S}\}  \hfill}  ({\bm{W}_S}(i,j)),\;\;\;\bm{W}_{hp}^S = {\bm{W}^S} - \bm{W}_{ne}^S \hfill \cr 
			\bm{W}_{ne}^T = \mathop {{\text{Avg}}}\limits_{\scriptstyle {\;\;\text{Min}} - \tau \%  \hfill \atop 
				\scriptstyle i,j \in \{ {(m,n)_T}\}  \hfill}  ({\bm{W}_T}(i,j)),\;\;\;\bm{W}_{hp}^T = {\bm{W}^T} - \bm{W}_{ne}^T \hfill \cr 
		\end{matrix}  \right.
		\label{eq:hp_neo}
	\end{equation}
	where the $\rm {Avg} (\cdot)$, which averages the learnable weights in the set, can be viewed as a smooth strategy to ensure the generality and smoothness of transferable neocortex structures. And $A-B$ denotes as  the complementary set of $B$ to set $A$. $\bm{W}_{ne}^S, \bm{W}_{ne}^T$ are decoupled neocortex neural  structures on respective spatial  and temporal blocks while $\bm{W}_{hp}^S, \bm{W}_{hp}^T$ are hippocampus neural structures for two blocks. 
	Comparing with existing OOD generalization such as CauSTG (requiring multiple training over different environments), the efficiency and adaptions of our proposed neural disentanglement lies in that we do not require too much memory, but only update the ${\bm{S}_{sp\Delta}^{tb}},{\bm{S}_{tp\Delta}^{tb}}$ every training unit, thus the disentanglement can be implemented along the usual  training process. 
	
		\begin{algorithm}[h]  
		\caption{Training procedure of ComS2T}  
		\label{alg:trainframe}  
		\begin{algorithmic}[1]  
			\Require  
			Main observations $\bm{X}$, Observed environment description ${\bf{E}}$;
			\Ensure 
			Neocortex structure on spatial and temporal blocks $ \bm{W}^{S}_{ne}, \bm{W}^{T}_{ne}$;  hippocampus structure on spatial and temporal blocks $\bm{W}^{S}_{hp}, \bm{W}^{T}_{hp}$;
			\For {iteration = 1, 2, ..., Q}
			\State Neural disentanglement decouples the learning neural spaces into initial  neocortex structure $\bm{W}^{S}_{ne}, \bm{W}^{T}_{ne}$  and hippocampus structure $\bm{W}^{S}_{hp},\bm{W}^{T}_{hp} $ based on Eq.~(\ref{eq:DeltaM}) to Eq.~(\ref{eq:hp_neo}). 
			\State Self-supervised prompt training based on Eq.~(\ref{eq:distrimu}) to Eq.~(\ref{eq:selfloss}) to obtain $\bm{P}_S, \bm{P}_T$.
			\State Fine-tune hippocampus structure with prompts  $\bm{P}_S, \bm{P}_T$ based on Eq.~(\ref{eq:SpatialFine}) to Eq.~(\ref{eq:TempFine}), and achieve updated $\bm{W}^{S}_{hp},\bm{W}^{T}_{hp} $.
			\EndFor
		\end{algorithmic}  
	\end{algorithm}
	
	\subsection{Self-supervised spatial-temporal prompt learning}
	\label{sec:sslprom}  
	Complementary learning system (CLS) is proposed to enable adaptive model evolution with data distribution. To this end, two important issues are identified in CLS, i.e., 1) how to endow the model with capacity of data adaptation, especially exploiting non-labeled samples for model evolution, 2) In a memory system, a brief but powerful summary can help remember better, thus how to design brief and informative prompts to guide the update of  model  becomes important.
	
	In this section, we design a pre-training strategy to achieve  continuous prompt representations in a self-supervised manner, thus we can take spatial-temporal prompts as an intermediate variable to deliver the variation of data   to main models. First, we respectively select informative spatial and temporal signals as the basic elements of prompts. We take longitude, latitude, location index as basic spatial information  i.e., 
	${\bm{e}_s}(i) = [lat,long;loc\_no]_{{v_i}} \in {\mathbb{R}^{2 \times E}}$, while consider day of week, time step, and time-series trend as representative temporal signal, i.e., $\bm{e}_t(t) = [Dw,Ts;Tr]_t \in {\mathbb{R}^{2 \times E}}$, where ';' denotes the division of line in the matrix.
	Second, we can explicitly model the distribution as a data summary over sequential observations, and construct a question-answer pair between spatial-temporal prompts and data summary to empower prompts sensitive to data distribution. Given the spatial location ${\bm{e}_s}(i)$ and temporal step $\bm{e}_t(t)$ at node $i$ and step $t$ in the spatiotemporal graphs, we model the continuous $\kappa$-step observations at corresponding spatial-temporal context as an observed distribution by corresponding parameters $(\mu_i^t,\sigma_i^t) \sim   X_i^{t:t + \kappa}$. Then we can easily regress these parameters through a carefully designed learning blocks. 
	To explore the relations between data distribution and spatial-temporal context, we construct a Spatial-Temporal Interaction Module (STIM) $g(\cdot)$ to capture the interactions between spatial and temporal contexts. It consists of a Compressed Interaction Network and an MLP structure, which allows field-level interactions between spatial and temporal prompts via inheriting the property of Deep Factorization Machine~\cite{guo2017deepfm}. With STIM, our self-supervised pre-training on prompts can be described as a regression task, 
	\begin{equation}
		({\widehat\mu _i},{\widehat\sigma _i})_t = g({\rm MLP}({\bm{e}_s}(i);{\bm{W}_{ps}}) \odot {\rm MLP}({\bm{e}_t}(t);{\bm{W}_{pt}});{\bm{W}_P})
		\label{eq:distrimu}
	\end{equation}
	where ${\bm{e}_s}(i) \in \bm{E}_s$ and $\bm{e}_t(t) \in \bm{E}_t$ are basic elements of spatial and temporal signals for construction of prompts, also accounting for the environment signals in our Fundamental Assumptions. 
	Through the regression of  parameters over  sequential observations, the intermediate representations of corresponding  spatial-temporal signals are taken as prompts respectively, i.e.,
	\begin{equation}
		\bm{P}_S = {\rm MLP}({\bm{e}_s}(i);{\bm{W}_{ps}}), \;\;
		\bm{P}_T = {\rm MLP}({\bm{e}_t}(t);{ \bm{W}_{pt}}) 
		\label{eq:prmpt}
	\end{equation}
	The above $\bm{P}_S, \bm{P}_T$ become the well-learned spatial and temporal prompts by imposing the following self-supervised learning objective,
	\begin{equation}
		Loss_{\mathit{self}}=\min \Sigma^{N}_{i} \Sigma^{T}_{i} (((\hat{\mu_{i} } )_{t}-(\mu_{i} )_{t})^{2}+(\left( \hat{\sigma_{i} } \right)_{t}  -(\sigma_{i} )_{t})^{2})
		\label{eq:selfloss}
	\end{equation}
	
	The above distribution can guide fine-tune on spatial and temporal prompts in a flexible manner upon accessing a few observations, regardless of the learning phases of training or testing. Meanwhile, such adjustment on prompts can deliver dynamics to hippocampus structure of our main model, thus empowering generalization to OOD scenarios and endowing it evolution capacity.

	\subsection{Progressive spatiotemporal learning}
	\label{sec:progressive}
	In this section, we  couple the prompt learning with a two-stage training, consisting of both warm-up and fine-tune, progressively achieving the evolution capacity. Concretely, for the whole architecture of ComS2T, it is composed of four stages, spatiotemporal model warm up and invariance decoupling, self-supervised pre-training, prompt-based fine-tune during model training, and test-time adaptation for data-adaptive testing.

	\textbf{Spatiotemporal model warm up and invariance decoupling.} Following Sec. ~\ref{sec:invardis}, we train spatial and temporal blocks with pair-wise main observations $\{(\bm{X},\bm{Y})\}$, namely model warm up, until achieving  stability of learnable parameters, i.e., the variations of errors become stable. We will activate the  efficient neural  disentanglement at the end of warm-up stage. We characterize the model behavior by retrieving  the accumulated variations of learnable weights at the stopped training unit $tb$, i.e., ${\bm{S}_{sp\Delta}^{tb}},{\bm{S}_{tp\Delta}^{tb}}$.  The neurons with $\tau \%$ smallest values in accumulated variations are disentangled as stable neocortex structure, while the complementary neuron set are classified as hippocampus structure. We can easily obtain the neural learning division, in spatial perspective $\bm{W}_{ne}^S, \bm{W} _{hp}^S$, and temporal perspective $\bm{W}_{ne}^T, \bm{W} _{hp}^T$. With all the parameters learned during warm up, $\bm{W}_s$ and $\bm{W}_T$ can be the initializations for following fine-tune process and enable the model preliminarily adapt to spatial-temporal observations. 
	
	\textbf{Self-supervised pre-training.} We then construct pair-wise training samples and exploit the distribution reconstruction to learn informative semantic spatial-temporal prompts by implementing Sec.\ref{sec:sslprom}. This allows the prompt learned by only accessing the potential distribution over main observations without sequential annotations (labels).
	
	\textbf{ST prompt-based fine-tune.} This stage allows the prompt as additional inputs of our framework with fine-tuning process. 
	We leverage the neocortex to preserve stable weights for transferring invariant relations across environments, and take spatial-temporal prompts to guide hippocampus update with distribution shifts where the meta information on spatial and temporal aspects can mostly reflect the data changes. 
	To facilitate the gradient propagation and semantic information aggregation, we inject the spatial and temporal prompts separately into the hippocampus structures regarding respective spatial and temporal learning blocks with careful dimension alignment. Specifically, we first freeze the neurons within neocortex structure, and integrate well-learned prompts with main observations as input to hippocampus structure. 
	Given the well-learned spatial prompt $\bm{P}_s$ and temporal prompt $\bm{P}_T$, in our fine-tune stage, the input of spatial learning block becomes,
	\begin{equation}
		{\bm{X}_{in}} = \bm{X} \oplus (\bm{W}_{ps}^{al}*{\bm{P}_s})
		\label{eq:SpatialFine}
	\end{equation}
	where $\oplus$ indicates element-wise addition. $\bm{W}_{ps}^{al}*{\bm{P}_s}$ accounts for dimension alignment between spatial prompt and main observations. 
	After that, we freeze neocortex structure and let hippocampus update. The output of spatial blocks can be written as,
	\begin{equation}
		{\bm{X}_S} = {F_S}({\bm{X}_{in}};\bm{W}_{hp}^S|\bm{W}_{ne}^S)
	\end{equation}
	Similarly, we update the input of temporal learning block by pre-alignment,
	\begin{equation}
		{\bm{X}_{ST}} = {\bm{X}_S} \oplus (\bm{W}_{pt}^{al}*{\bm{P}_T})
	\end{equation}
	Then the output of the temporal blocks can be written as,
	\begin{equation}
		\bm{Y} = {\Gamma _T}({\bm{X}_{int}}; \bm{W}_{hp}^T, \bm{W}_{ne}^T)	
		\label{eq:TempFine}
	\end{equation}
	
	With our complementary learning strategy, the neocortex structure transfers stable relations while introduces prompt signals derived from external factors to update hippocampus structures, allowing the model to be  dynamic with data distribution changing over time. Our design innovatively fixes invariant relations within spatial-temporal observation, and regress the residual observations with contexts, which enables fine-tune only requiring limited computations.
	To this end, the whole architecture integrates spatial and temporal learning blocks where each block consists of corresponding neocortex and hippocampus structures.	Formally, the integrated model becomes,
	\begin{equation}
		\bm{Y} = f^*(\bm{X},{\bm{P}}_S,{\bm{P}}_T;(\bm{W}^{S}_{hp}||\bm{W}^{S}_{ne})||( \bm{W}{{}^{T}_{hp}{}} ||\bm{W}^{T}_{ne}))
		\label{eq:finaloutput}
	\end{equation}
	Symbol $||$ is the concatenation operation between neuron structures.  In addition, to follow up the latest findings in complementary learning~\cite{kumaran2016learning}, i.e.,  neocortex can slowly update to accommodate the new knowledge and dynamic hippocampus structure, we can further devise an alternate learning strategy to update them with different frequencies, i.e., the neocortex regularly update  every $K$ batches while hippocampus structure updates every batch. Since it is not the main issue of our complementary learning, we leave it as our future work.
	
	\textbf{Test-time adaptation.} To enable a true evolvable model, we take advantage of self-supervised learning to fine-tune the prompts during test time, which enables the efficient update on partial model parameters. When new observations arrive, we can sample a small number of batches of observations with their counterpart spatial and temporal descriptions, $\widetilde{\bm{X}}_{\mathit{test}} \sim \bm{X}_{\mathit{test}}$. Then we can fine-tune the prompt representation $\bm{W}_{ps}$ and $\bm{W}_{pt}$ with new data distribution, and update the intermediate embedding $\widetilde{\bm{P}}_S, \widetilde{\bm{P}}_T$ based on Eq.~(\ref{eq:distrimu}) and Eq.~(\ref{eq:selfloss}). Finally, we can correspondingly obtain the new learning outputs by re-exploiting  Eq.~(\ref{eq:finaloutput}),
	\begin{equation}
		\bm{Y}_{\mathit{test}}  = f^*(\bm{X}_{\mathit{test}},\widetilde{\bm{P}}_S, \widetilde{\bm{P}}_T;(\bm{W}^{S}_{hp}||\bm{W}^{S}_{ne})||( \bm{W}{{}^{T}_{hp}{}} ||\bm{W}^{T}_{ne}))
		\label{eq:testpred}
	\end{equation}
		
	\begin{algorithm}[h]  
		\caption{Testing procedure of ComS2T}  
		\label{alg:testframe}  
		\begin{algorithmic}[1]  
			\Require  
			Main testing observations $\bm{X}_{\mathit{test}}$, Observed environment description ${\bf{E}}_{\mathit{test}}$, Well-learned  parameters of ComS2T $\bm{W}^{S}_{ne}, \bm{W}^{T}_{ne}, \bm{W}^{S}_{hp}, \bm{W}^{T}_{hp}$; 
			\Ensure 
			Prediction results $\widehat{\bm{Y}}_{\mathit{test}}$;
			\State Sample partial observations from testing set $\widetilde{X} \sim  \bm{X}_{\mathit{test}}$ and compute the parameterized distribution $(\widetilde{\mu}, \widetilde{\sigma})$; 
			\State Update $\bm{P}_S, \bm{P}_T$ into $\widetilde{\bm{P}}_S, \widetilde{\bm{P}}_T$ based on Eq.~(\ref{eq:distrimu}) to Eq.~(\ref{eq:selfloss}); 
			\State Implement prediction based on Eq.~(\ref{eq:testpred}) and output $\widehat{\bm{Y}}_{\mathit{test}}$.
		\end{algorithmic}  
	\end{algorithm}
	\textbf{Learning objective.} We take Mean Absolute Error as the main learning objective for our training architecture in both warm-up and fine-tune stages. For self-supervised learning, we exploit the  reconstruction loss $Loss_{\mathit{self}}$  as the objective. 
	
	\textbf{Model efficiency analysis.} We further provide a brief analysis to support the satisfactory efficiency of our ComS2T. Given the OOD scenario, we compare the computation loads, indicated as the 
	updated parameter number with gradient backpropagation, against comparable baseline solutions including CauSTG, PECPM, and TrafficStream.  Assuming each model is equipped with $L$ parameters, the streaming data will experience $P$ times of  distribution shifts. The number of parameters regarding prompt update can be denoted as $ |\bm{W}_{ps}| +|\bm{W}_{pt}|+  |\bm{W}_P| = E_P$.  And there are $K$ temporal environments, above three models will update $\gamma\%$ of parameters, where  $\gamma\% = 1-\tau\%$ in our ComS2T. Specifically, CauSTG learns $K$ submodels across temporal environments  and then update model during testing scenario. PECPM and TrafficStream will update the $\gamma\%$  parameters of the whole model.  For PECPM and TrafficStream, PECPM updates partial parameters of model when the distribution changes while TrafficStream takes an experience-reply strategy  for model update. In contrast, our ComS2T  only  undergoes two stages, i.e., warm-up learning and prompt-based fine-tune,  then only a few parameters of spatial-temporal prompts will update when distribution changes. The number of updated model parameters have been listed in Table.~\ref{tab:effiency}. Obviously, with the increases of data distribution shifts and road network expansions, the dimensions for both spatial and temporal prompts should be update, i.e.,
	$L*\gamma\% + P\times E_P  = L\gamma\% +PE_{P}  \ll LP \times\gamma\%$, thus only updating prompts with fewer parameters can open a new avenue towards OOD learning  and streaming data processing. Therefore, ComS2T can exactly enjoy the superiority of once training for all updates.

	\begin{table}[]
		\centering
		\caption{Model efficiency comparisons across baselines.}
		\label{tab:effiency}
		\begin{tabular}{l|c}
			\toprule
			\hline
			\tabincell{l}{Model} & {Number of parameter updates under temporal shift}   \\ \hline
			CauSTG  & $KL+ LP*\gamma\%$       \\ \hline
			PECPM   & $L+ LP*\gamma\%$      \\ \hline
			TrafficStream &  $L+ LP*\gamma\%$      \\ \hline
			ComS2T     & $L+  L*\gamma\%+ P*E_P  $  \\ \hline     
			\bottomrule                        
		\end{tabular}
	\end{table}
	
	\textbf{Summary.} It is worth noting that we only explore the dynamics of learnable prompts to enable the model update efficiently.
	Our ComS2T is efficient and reliable, its efficiency lies in its potential of decoupling associations during training, and the use of prompt  modeling to obtain a continuous mapping relationship between the context environment and  real distribution over main observations.  It allows delivering the distribution change from observation to prompt and subsequently the hippocampus structures of ComS2T during both training and testing stages. The learning and testing procedures of our ComS2T are respectively described in Algorithm.~\ref{alg:trainframe}  and ~\ref{alg:testframe}, and detailed experiment settings can be found in Sec.~\ref{sec:implemtdetail}.

	%
	%
	%

	\section{Experiment}
	We collect four types of spatial-temporal data and design  various learning-testing scenarios to imitate the distribution shifts on both spatial and temporal aspects.
	
	
	\subsection{Dataset description}
	We take different categories of  spatiotemporal datasets such as traffics, air quality and smart grids,  for verification of our data-adaptive learning architecture. The statistics of datasets can be found in Table.~\ref{tab:dataset}.
	

	\begin{itemize}
		\item  \textbf{{SIP (Traffic)}:} It is the camera surveillance capturing traffic volumes in Suzhou Industry Park (SIP), Suzhou. 
		\item {\textbf{Metr-LA (Traffic)}:} Traffic attributes such as speed, detected by highway loop detectors of Los Angeles, USA. We reach this dataset via literature~\cite{li2018diffusion}.
		\item  \textbf{{KnowAir (Air quality)}:} PM2.5 concentrations, covering 184 main cities of China~\cite{wang2020pm2}. 
		\item  \textbf{{Temperature (Climate)}:} Urban numerical temperature  covering the same 184 cities as KnowAir~\cite{wang2020pm2}. 
	\end{itemize}
	
	\begin{table}[]
		\centering
		\caption{Dataset  statistics}
		\small
		\begin{tabular}{lllll}
			\toprule
			\hline
			Dataset     & \tabincell{l}{Node \\ \# } & \tabincell{c}{  Time\\ step \#} & \tabincell{c}{Time \\ span}    & \tabincell{l}{Interval \\length} \\ \hline
			SIP    & 108         & 25,920              & \tabincell{l}{01/01/2017- \\ 03/31/2017 }  & 5min           \\
			Metr-LA     & 207         & 34,272              & \tabincell{l}{03/01/2012-\\ 06/30/2012} & 5min             \\
			KnowAir     & 184         & 11,688              & \tabincell{l}{01/01/2015- \\ 12/31/2018 } & 3h      \\
			Temperature & 184         & 11,688              & \tabincell{l}{01/01/2015- \\ 12/31/2018 } & 3h              \\ \hline
			\bottomrule
			\label{tab:dataset}
		\end{tabular}
	\end{table}

	\subsection{Learning with spatial-temporal OOD settings}
	\label{sec:OODset}
\begin{figure*}[ht]	
		\centering
		\includegraphics[width=0.9\linewidth]{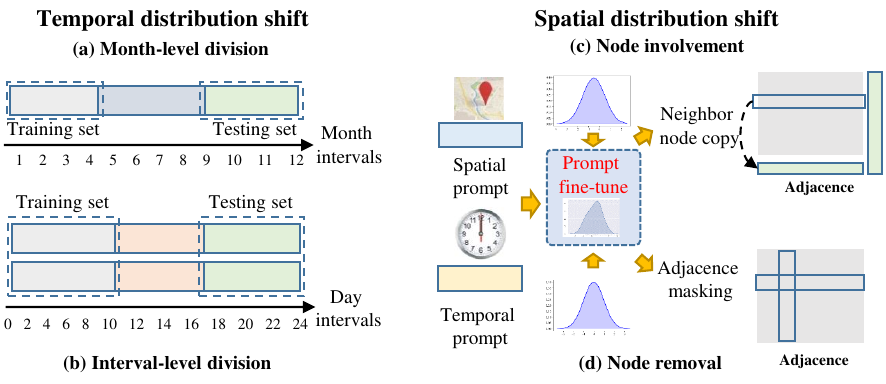}
		\caption{Experimental settings of four OOD scenarios}
		\label{fig:exp01}
	\end{figure*}

	We construct data distribution shifts on temporal perspective while imitate structure shifts on spatial aspect, as illustrated in Fig.~\ref{fig:exp01}. First, \textbf{temporal distribution shift} can be imitated by two training-testing divisions according to data distribution characteristics over different datasets.
	\begin{itemize}
		\item Interval-level division: For traffic datasets of \textbf{SIP} and \textbf{Metr-LA} those are highly dynamic, it is observed that evolution patterns on two half days are totally different, we can well imitate the temporal distribution shift. We thus organize training sets by collecting all the same day intervals (e.g., every 8:00-16:00) for model learning, while perform testing on the other unseen day intervals (e.g., every 1:00-7:00).   
		\item  Month-level division:   For air quality and climate datasets those are relatively static within short term but can be varied seasonally, we  divide the whole-year records into four trimesters, where we train with the two trimester while test on one of the trimester.  
	\end{itemize}
	
	Second, \textbf{spatial distribution shifts} are  realized by the involvement of new nodes and  removal of existing nodes. 
	\begin{itemize}
		\item  Node involvement: We actively mask a series of existing nodes during training and add them back during testing stage to simulate the new connections of the graph structure. 
		\item Node removal: Similarly, we  remove some existing nodes during  testing stage for imitating the node disappearance in the dynamic graph  structure.  
	\end{itemize}
	
	\subsection{Implementation details}
	\label{sec:implemtdetail}
	\textbf{Dataset organization and processing.} For each dataset, we organize them as sample groups following settings in Sec.~\ref{sec:OODset}. For \textbf{SIP and Metr-LA}, we take 1/3 of samples for training, i.e., every 8:00-16:00, samples within every 16:00-24:00 for validation, samples within every 0:00-7:00 for testing where 0:00-1:00  for test-time data adaptation.  
	For \textbf{air quality and climate datasets}, we take samples during every first six months (January to June) for training,  samples during July and August for validation,  samples during October to December for testing where samples of September is considered for test-time data adaptation. We follow this division for each baseline to ensure the  fairness of comparison.
	
	Regarding data processing, we encode the categorical context with one-hot embedding and transfer them into fixed-length vectors. Our target is to construct the data-adaptive  model to predict next 12 slots based on the current 12 frames ($\tau=12$) under OOD settings including both temporal and structural shifts.  
	
	\textbf{Deep learning implementation.}
	For general settings, all the methods are implemented using  Pytorch 1.10.0 and evaluated on one Tesla V100 GPU. To guarantee fair comparison, we perform grid search to tune the hyperparameters for all baselines over three datasets.  The hyperparameter configurations of our ComS2T can be found in Table~\ref{tab:config}.
	
	\begin{table}[]
		\small
		\caption{Configuration of ComS2T. The configurations are displayed in the order of SIP, Metr-LA, KnowAir and Temperature if values  should be specified by datasets. }  
		\begin{tabular}{ll}
			\bottomrule
			\textbf{Parameter}            & \textbf{Concrete values}    \\  \hline 
			Backbone of ComS2T    & GraphWaveNet (GWN)	\\ 
			Learning rate  & 1e-4    \\
			Dimension of spatial-temporal prompts & (64,16,16,32) \\
			Percentage of stable neocortex $\tau$ & (60\%, 60\%, 60\%,70\%)\\
			Hidden dimension of GNNs  &  32   \\ 
			TCN kernel dimension &  (12,6,3) \\
			Batch size  & 64 \\ 
			Optimizer     & Adam    \\ 
			\bottomrule
		\end{tabular} 
		\label{tab:config}  
	\end{table}
	
	\begin{table*}[]
		\small
		\caption{Performance comparisons on OOD scenarios against baselines (Metric: MAE). The \textbf{best results} are bold while \underline{second best} are underlined. }
		\label{tab:compar-performance}
		\begin{tabular}{l|ccc|ccc|ccc|ccc}
			\toprule
			\hline
			& \multicolumn{3}{c|}{SIP}  & \multicolumn{3}{c|}{Metr-LA}  & \multicolumn{3}{c|}{KnowAir} & \multicolumn{3}{c}{Temperature} \\ \hline
			& \tabincell{c}{Temp \\shift}   & \tabincell{c}{Node \\ involve}  & \tabincell{c}{Node \\ removal} &  \tabincell{c}{Temp \\ shift}   & \tabincell{c}{Node \\ involve}  & \tabincell{c}{Node \\ removal} & \tabincell{c}{Temp \\ shift}   & \tabincell{c}{Node \\ involve}  & \tabincell{c}{Node \\ removal}& \tabincell{c}{Temp \\ shift}   & \tabincell{c}{Node \\ involve}  & \tabincell{c}{Node \\ removal}  \\ \hline
			MTGNN         &        44.25 &	49.80 &	\underline{46.28}	& 6.86 &	6.26 &	6.85 &	\underline{36.73} &	46.38 &	39.56 &		\underline{6.87} &	7.22 &	\underline{6.97} \\	\hline
			GWN   &    44.17 &	50.76 & 	47.57 & 	6.16	& 5.52 & 	7.64 & 	38.19 & 	36.48 & 	40.05 & 	7.47 & 	8.43 & 	7.92         \\	\hline
			ST-SSL   &    44.55 & 	45.23 & 	46.55 & 	6.58 & 	5.76 & 	7.77 &  	39.17 & 	38.72 & 	43.09 & 	8.66 & 	8.17 & 	7.95   \\	\hline
			CauSTG    &   	\underline{43.47} 	&  50.10 &	46.70 &	5.97 &	4.68   & 	6.82 &	37.86 &	35.75 &	39.32 &	7.07 & 	7.44 & 7.59 \\	\hline
			CaST & 43.52    &	48.67 &	{46.54} &	5.59  &	4.98 & 7.46 &	38.88 &	\textbf{33.42}	& \underline{38.35} & 7.35 &	\underline{7.16} & 7.04 \\\hline
			IRM+GWN    & 43.73 & 	47.43  &  	47.14 &	\underline{5.53}  & 4.97  &  7.47  & 	39.50  &  	35.62 &	39.61 &	7.33  &  	8.28 &  	7.22 \\	\hline
			PECPM & 44.78 &	\underline{43.51} &	47.78&	6.18& \underline{4.65} &	\underline{6.80} &	38.94&	36.27&	39.53&	7.56&	7.25&	7.11 \\ \hline
			TrafficStream	& 45.67 & 45.32&	47.95	& 6.47 &	4.86 &	6.92&	39.15&	37.03&	40.48&	7.73&	8.22&	7.90 \\ \hline
			ComS2T        &   \textbf{41.12}    &   \textbf{39.88}  &	\textbf{44.52} &	\textbf{4.38}	& \textbf{4.35} &	\textbf{5.62} & \textbf{35.32}  & 	\underline{34.48} & \textbf{38.21} &	\textbf{6.82} & \textbf{6.90} & \textbf{6.88}           \\  \hline
			\tabincell{l}{Beyond \\ second best}   & 5.41\%  &	8.36\% &	3.79\%  &	20.70\% &	6.40\% &	17.30\% &	3.83\% &	-3.17\% & 0.36\% &	0.73\% &	3.63\% &	1.19\%\\ \hline
			\bottomrule
		\end{tabular}
	\end{table*}
	\textbf{Prediction details under OOD settings.} 
	The training process of our ComS2T can be three-fold, self-supervised pre-training for spatial-temporal prompts, model warm up and hippocampus structure disentanglement, and prompt-incorporated model fine-tune. With the placeholders of prompts, we can deliver  conditional prompts forward to main learning structure, then the prompts and hippocampus structures will be jointly fine-tuned to allow the increase of generalization capacity during fine-tune stage. 
	During testing stages under temporal shifts, we exploit the temporally nearest samples for test-time model adaptation, which allows the update of prompts conditional on distribution shifts. 
	
	For structure shifts, when a new node is introduced, we exploit the distribution-supervised learning scheme to update the spatial and temporal prompts based on a few new observations. For adjacent matrix, we impose a node copy strategy ~\cite{zhou2023maintain}, which finds the most proximal nodes to new ones and copies the  adjacencies of existing similar nodes to new ones, thus constructing an extended relational spatial adjacency for testing.  When an existing node is removed from the spatiotemporal graph, we re-train spatial and temporal prompts, and mask the corresponding line and column of the removed nodes in adjacent matrix for dimension alignment.  

	\textbf{Evaluation metrics.} Each baseline and our neural ComS2T are implemented five times and the averaged errors are reported. We take Mean Absolute Error (MAE) as the main metric for evaluation. The error can be written as, i.e.,
	\begin{equation}
		{\text{MAE}} = \frac{1}{{TN}}\sum\limits_{t = 1}^T {\sum\limits_{j = 1}^N {|y_i^t - } } \widehat{y}_i^t|
	\end{equation}
	where $\widehat{y}_i^t$ is the predicted observation of node $i$ at time step $t$, while ${y}_i^t$ is corresponding ground-truth.

	\subsection{Baseline}
	Our baselines  are three-fold, including three satisfactory ST learners,  two causal-based ST learners, and two Continuous learning-based ST learners.
	\begin{itemize}
		\item \textbf{MTGNN}: A graph-based multi-variate time series learning without  defining explicit graph topology~\cite{wu2020connecting} (\underline{ST learner}).  	 
		\item \textbf{GraphWaveNet (GWN)}: A graph-based traffic prediction model that integrates TCNs and  GCNs~\cite{wu2019graph} (\underline{ST learner}).
		\item \textbf{ST-SSL}: A State-of-the-Art  learning architecture explicitly considering discrimination on  spatial and temporal  dimensions~\cite{ji2023spatio} (\underline{ST learner}).
		\item \textbf{CauSTG}: An emerging Causal-based invariant learning for spatial-temporal data~\cite{zhou2023maintain} (\underline{Causal learner}).
		\item \textbf{CaST}:  A causal lens spatial-temporal learning framework, which explicitly models the environments and imposes the backdoor adjustment~\cite{xia2023deciphering} (\underline{Causal learner}).  
		\item \textbf{IRM+GWN}: We especially integrate the invariant risk minimization with GraphWaveNet to test the generality of its framework  (\underline{Causal learner}). 
		\item \textbf{PECPM}: A memory-based  continuous learning via pattern expansion along urban expansion~\cite{wang2023pattern} (\underline{Continuous learner}).
		\item \textbf{TrafficStream}: An experience reply-based  continuous learning framework  for traffic  flow prediction~\cite{chen2021trafficstream} (\underline{Continuous learner}).
	\end{itemize}

	\subsection{Analysis of performances against competitors}
	The comprehensive experimental comparisons are shown in Table.~\ref{tab:compar-performance}. Note that the temporal shift, node-level involvement and node-level removal are abbreviated as 'Temp shift, Node involve and Node removal' in our result tables. According to the characteristics of datasets, we take interval-level division to imitate the temporal shift on Traffic datasets (SIP and Metr-LA) while take month-level division for air quality and climate datasets (KnowAir and Temperature). The improvements beyond second best baseline are illustrated at the bottom of Table.~\ref{tab:compar-performance}. Overall, our ComS2T achieves  consistent superior performances against baselines under most scenarios, improving the performances from 0.73\% to 20.70\% under temporal distribution shifts, while promoting  1.19\% to 17.30\% under structural shifts.  More specifically, it shows a significant improvement on Metr-LA dataset, and this may be attributed to the nice regularity between spatial-temporal prompts and main observations.  The detailed four observations on respective categories of solutions are elaborated as follows. 
	
	
	\textbf{Obs1. Comparison against traditional ST learners.} Although traditional ST learners reveal satisfactory performances on settings of consecutive sequence forecasting, but they still fall short under distribution shifts, especially on two traffic datasets. Promisingly, MTGNN
	and ST-SSL reveal some robustness to structral shifts. It is mostly because that the learnable adjacencies  are well-transferred to new nodes with node copy strategy while the step-wise and node-wise self-supervised signals may play vital role in obtaining distinguished patterns for generalization. Thus, the potential advantage of SSL learning lies in  improving  representation capacity with self supervision, which is also inherited into our ComS2T. 
	
	\textbf{Obs2. Comparison with ST model with invariant learning.} Some pioneering models have taken invariance and transferability across environments into consideration to counteract the temporal distribution shifts, e.g., CauSTG, and IRM+GWN. The empirical results show that they can exactly improve the OOD learning capacity  but they are still inferior to our ComS2T. The underlying reason can lie in that they only transfer the invariance to OOD scenarios while  no specific solutions for model update and data adaptation. These two methods reasonably trap into suboptimal performances.

	\textbf{Obs3. Comparison with ST continuous learning.} Further, for those prediction models explicitly considering environment variations, e.g., CauSTG, CaST, and TrafficStream, which either exploits the closed environment division ~\cite{zhou2023maintain} and codebooks~\cite{xia2023deciphering}, or employs experience reply to re-train the model~\cite{chen2021trafficstream}, they still fail to fully exploit the available environment information to improve the adaptation capacity. In contrast, our ComS2T leverages both advantages of self-supervised prompts and complementary learning to accommodate spatial and temporal prompts by establishing bridges between main observations and environment prompts, contributing to at least 8.17\%  improvement under temporal shifts and 3.16\%  improvement under structural shifts against the typical continuous learning scheme PECPM.

	\textbf{Obs4. Comparison under structural shifts.} Finally, even though CauSTG has considered the spatial shifts and PECPM focuses on the issue of road network expansion and achieves several second-best results under structural shifts, it is still  empirically inferior to our ComS2T. Our work explicitly involves  spatial structural contexts by updating spatial prompts with new observations. It is observed that our solution can significantly outperform both CauSTG and PECPM, for instance, it accounts for the improvement of 3.01\% against CauSTG under temporal shift of SIP, and 17.30\% promption against PECPM under node removal of Metr-LA. Besides, these mentioned  solutions still suffer the efficiency issue for multiple training of submodels (CauSTG) and computation of pattern-level matching (PECPM). In our ComS2T, when the urban structure changes, we only require preliminarily update on spatial-temporal prompts with a few observations, and then it can be well-generalized on testing set with new structures, inclusion or exclusion of new nodes. 
	
	In summary, we can conclude that our ComS2T is superior to all other baselines on two aspects, i.e., 1) Without sacrificing the memory storage for new pattern preservation and computational burden on sequence-level pattern matching, our ComS2T directly disentangle the stable and dynamic neural architectures and actively update the neural networks in an overall pipeline, resulting in its superior efficiency. 2) Our ComS2T takes both advantage of self-supervised prompt with distribution reconstruction and the complementary learning architecture, which allows the flexible prompt update with new observations and realizing the exact data adaptation spatiotemporal learning framework model. 
	\begin{figure*}[ht]	
		\centering
		\includegraphics[width=\linewidth]{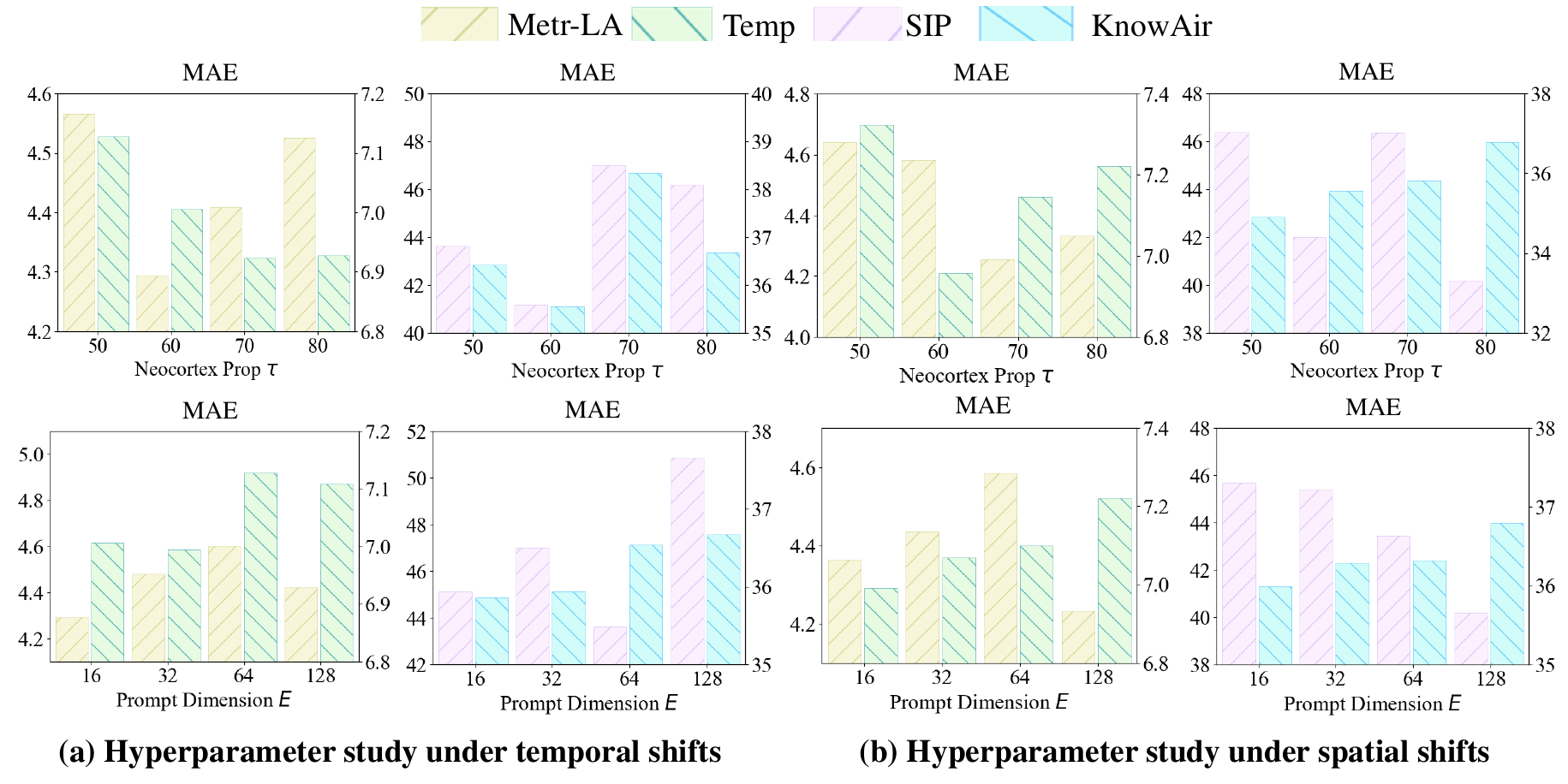}
		\caption{Hyperparameter study under both spatial and temporal shifts}
		\label{fig:hyper}
	\end{figure*}

	\begin{table}[]
		\tiny
		\caption{Performance comparisons on variants of ComS2T (Metric: MAE). The \textbf{best results} are bold and the \underline{second best} are underlined.}
		\label{tab:ablation}
		\begin{tabular}{l|cc|cc|cc|cc}
			\toprule
			\hline
			Datasets  & \multicolumn{2}{c|}{SIP}      &\multicolumn{2}{c|}{Metr-LA}      
			& \multicolumn{2}{c|}{KnowAir}    &\multicolumn{2}{c}{Temperature} \\ \hline
			\tabincell{l}{OOD  \\ scenarios} & \tabincell{c}{Temp \\shift}  & \tabincell{c}{Node \\involve} 	& \tabincell{c}{Temp \\shift}  & \tabincell{c}{Node \\involve}
			& \tabincell{c}{Temp \\shift}  & \tabincell{c}{Node \\involve}
			& \tabincell{c}{Temp \\shift}  & \tabincell{c}{Node \\involve} \\\hline
			Non-hip    & 54.48       & 44.62    & 5.48    & 5.10     & 36.47    & 38.08    & \underline{6.91}      & 7.05      \\ \hline
			Non-SSL    & 45.73    & \underline{40.47}      & \underline{4.82}     & 4.74   &36.46                              & 37.80     & 6.98    & 7.32    \\ \hline
			Non-Prompt & 49.21     & 46.47        & 4.99   & 4.59    &38.33   & 37.43   & 6.99  & 7.09  \\ \hline
			Non-TTF    & \underline{45.12}   & 44.11    & 4.83   & \underline{4.44} & \underline{35.14}    & \underline{36.56}  & 7.27      & \underline{6.92}   \\\hline
			ComS2T     & \textbf{41.12}  & \textbf{39.88}    &  \textbf{4.38}  & \textbf{4.35} & \textbf{35.32}                              & \textbf{34.48}       &  \textbf{6.82} & \textbf{6.90}     \\ \hline     
			\bottomrule                        
		\end{tabular}
	\end{table}
	
	\begin{figure*}[ht]	
		\centering
		\includegraphics[width=\linewidth]{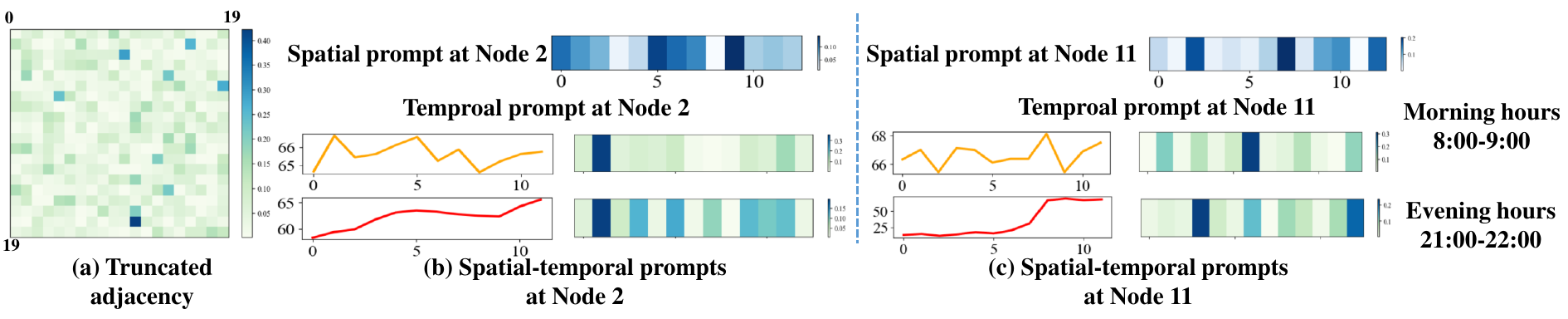}
		\caption{Visualization of spatial-temporal prompts under different contexts.}
		\label{fig:case-prom}
	\end{figure*}
	\subsection{Ablation study}
	We remove the specific modules in our ComS2T to verify the   contribution  of each well-designed component or learning strategy. We conduct experiments on two OOD scenarios of both temporal shifts and structural node-level shifts, where the results are illustrated in Table.~\ref{tab:ablation}.
	\begin{itemize}
		\item 	Non-hip: Do not identify the dynamic hippocampus structure and take the whole architecture for OOD inference. We train and update the whole neural architecture  with prompts but without explicitly identifying the hippocampus and neocortex structures,  which helps verify the effectiveness of hippocampus structure. 
		\item   Non-SSL: Remove the self-supervision signals  of learning spatial-temporal prompts, and take random initialization to replace the prompt training process.
		\item   Non-prompt: Update the hippocampus structure  of our model without  spatial-temporal prompts, just update it with pairwise $\{(\bm{X}, \bm{Y})\}$.
		\item  Non-TTF: Do not update the prompt during testing stage. 
	\end{itemize}
	
	As shown in results, our well-designed ComS2T clearly beat against four variants and achieves the best performances. Specifically, the performance experiences a prominent drop when the hippocampus structure is disabled on traffic datasets (SIP and Metr-LA),  verifying the exact  effectiveness of our complementary learning architecture. For KnowAir and Temperature, these two datasets are respectively sensitive to prompt description and test-stage adaptation, which are also served as two vital components for data adaptation. The reason for the heterogeneous sensitivity of components to datasets may be the differentiated characteristics of datasets, where traffics are with explicit immediate dynamics, while air quality and climate observations are more regular by considering the seasonal and location-based prompts.
	To conclude, even  performance variations  across datasets  are illustrated, the consistent drops in variants and the better performances with integrated ComS2T confirm the designs and intuitions of coupling complementary learning with spatiotemporal forecasting.

	\subsection{Hyperparameter analysis}
	To test the  sensitivity of our ComS2T, we select two crucial hyperparameters to observe how the  model behave along with the changes of parameters. Our experiments are conducted over temporal distribution shift on all datasets, and empirically optimized hyperparamters are listed as below,	 
	\begin{itemize}
		\item  Percentage accounting for stable neocortex $\tau$, we let it range from $\{50\%, 60\%, 70\%\, 80\%\}$. 
		\item The dimensions of spatial and temporal prompts, we consider them as the same dimension $E$,  and let it change within $\{16, 32, 64, 128\} $. 
	\end{itemize}
	The model performance variations can be found in Figure~\ref{fig:hyper}. For temporal shift scenarios,  Metr-LA and knowAir both achieve best at 60\% proportion of neocortex  with  the  prompt dimension of 16. SIP reaches its best with 60\% proportion of neocortex and  prompt dimension of 64, while Temperature  reaches its best with 70\% proportion of neocortex and dimension of 32. The higher stable proportion on Temperature suggests that city-level temperature is with higher regularity and stability than other urban attributes such as traffics and air quality  under temporal shifts. And the larger hidden dimension for SIP demonstrates the  potential dynamics of corresponding set  requires more fitting capacity of neural networks. 
	Regarding  structural shifts, four datasets respectively adapt the preserved stable ratios to 80\%, 70\%, 50\% and 60\% at the best performances.  For the dimension of prompts, we observe that SIP and Metr-LA both achieve the best performance at the hidden dimension of 128, while KnowAir and Temperature both perform best at the dimension of 16. This is because the dynamics of traffics requires more fitting capacity.  In summary, in this subsection, we can not only achieve satisfactory  results with our hyperparameter studies, but  provide insightful  urban analysis for further research on cities.

	\subsection{Detailed case study and model exploration}
	Our case studies are provided to answer the following two \textbf{R}esearch \textbf{Q}uestions (\textbf{RQ}), with empirical and visualized results on specific cases and detailed analysis.
	\begin{itemize}
		\item \textbf{RQ1:} How prompts interpret the dynamic spatial and temporal contexts, and whether the prompts can be adaptive with the changes over distribution of main observations? 
		\item \textbf{RQ2:} How the learnable parameters of main ComS2T architecture behave along with the learning process, whether the disentanglement and partial update of neural structures are effective for performance improvement and generalization.
	\end{itemize}
	
	\textbf{Visualization on prompts with distinctiveness.} First, we visualize the well-learned spatial and temporal prompts at node 2 and 11 on Metr-LA~\footnote{It is a dataset of traffic speed.} in Fig.~\ref{fig:case-prom},  accompany with corresponding sequential observations of traffic speed. The selected periods for visualization are  workday morning peak hours and evening hours after peaks. It is observed that the two sequences are distinctive on patterns where the former peak hours experience larger  speed variations with underlying high volumes across all steps while the latter one shows relatively stability with time goes on. The visualized results can deliver that the absolute differences on learned prompts, where the similar day intervals illustrate some similarity, e.g., similar series patterns are identified  during respective morning peak hour and evening hours after peaks. These visualized intermediate results with reasonable distinctiveness and similarity demonstrate that our self-supervised learning signal of distributions can effectively guide the optimization process of prompts and obtain distinguished prompts against different distributions, allowing  data adaptation to be delivered forward to update of hippocampus structures. 

		\begin{figure}[ht]	
		\centering
		\includegraphics[width=\linewidth]{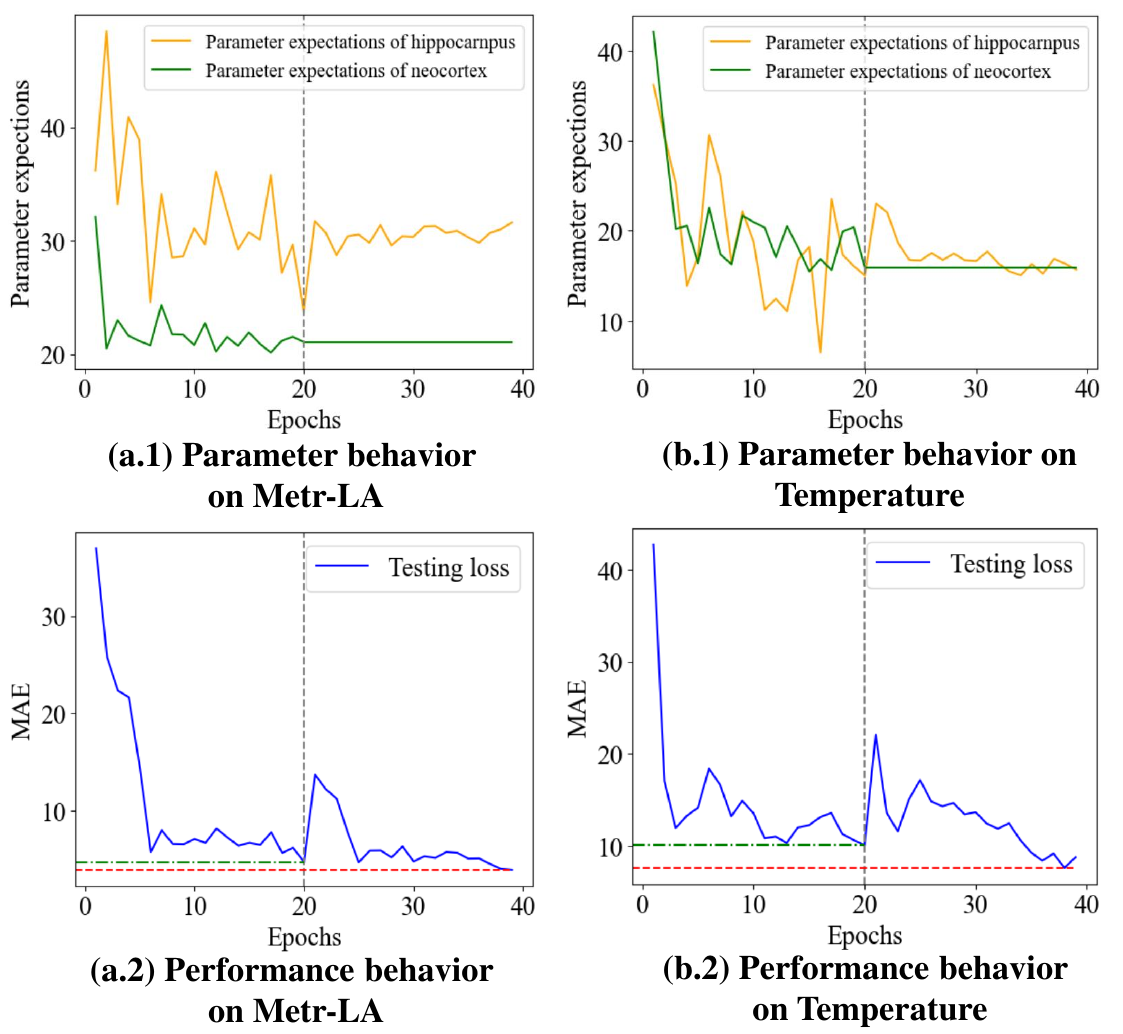}
		\caption{Learning behavior visualization of ComS2T.}
		\label{fig:learn-behave}
	\end{figure}
	
\begin{figure}[ht]	
		\centering 
		\includegraphics[width=\linewidth]{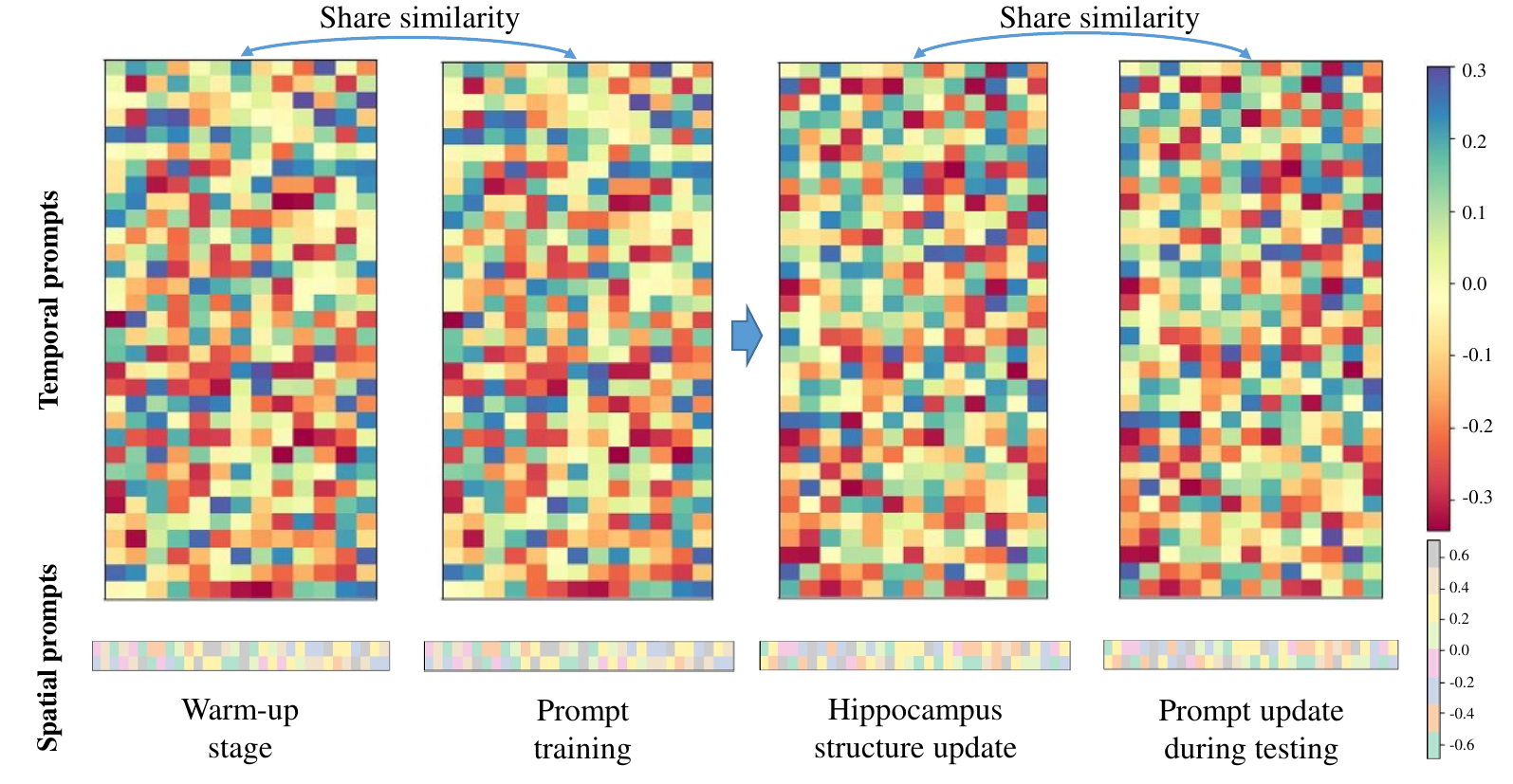}
		\caption{Fine-grained prompt visualization during learning stages on Metr-LA.}
		\label{fig:weightMetrLA}
	\end{figure}

\begin{figure}[ht]	
		\centering 
		\includegraphics[width=\linewidth]{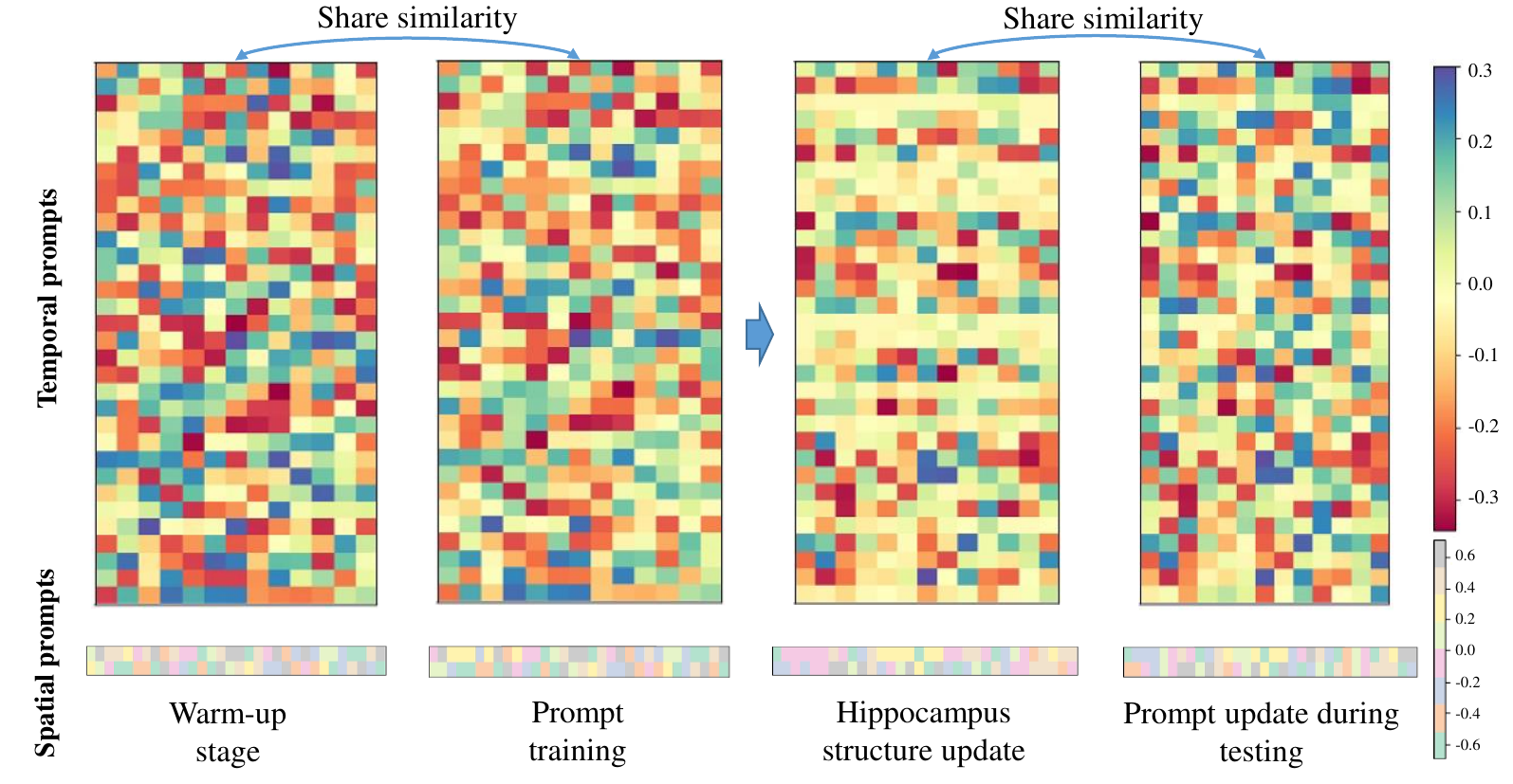}
		\caption{Fine-grained prompt visualization during learning stages on Temperature.}
		\label{fig:weightTemp}
	\end{figure}

	\textbf{Training process visualization.} Second, we illustrate the  parameter behaviors and corresponding performance variations along with the training procedure in Fig.~\ref{fig:learn-behave} to Fig.~\ref{fig:weightTemp} on  Metr-LA and Temperature. To be specific, the parameter behaviors are 
	demonstrated on two-fold, i.e., the collective behavior and individual behaviors. 1) We visualize \underline{the expectation of the parameters} over neocortex and hippocampus neural structures, as collective parameter behaviors, and take performance indicator of \underline{MAE errors}  along with the number of training epochs in Fig.~\ref{fig:learn-behave}. 
	We have marked the knee point of the beginning of prompt-conditioned hippocampus update with dashed line in our Figure. 2) To investigate how individual parameters behave,  we also introduce the  fine-grained visualizations of both spatial and temporal prompts during learning process in Fig.~\ref{fig:weightMetrLA} and Fig.~\ref{fig:weightTemp}. 	
	For collective parameter behaviors, it is observed that the hippocampus structure experiences a heavy fluctuation at the knee point while fluctuation slows down as the learning process continues. For performances, the errors first climb up and then drop down to reach the stability, where the increases of errors and fluctuation of parameters reflect the adaptation and adjustment process of neural architecture with incorporation of spatial-temporal prompts. With the progressive learning with informative prompt signals, the errors are decreasing and eventually outperforming the performance of warm-up stage. For individual ones, the prompt parameters show distinctiveness during each stage, where it also shows inter-stage similarity between warm-up and prompt training stages, as well as between hippocampus  update and testing stages. This  suggests that each update process can achieve data adaption and the hippocampus update can impose sufficient impacts for model optimization and data adaptation.
	
	The above first two cases respectively illustrate the distinctiveness of spatial-temporal prompt representation, and the effectiveness of our disentanglement and update process in ComS2T. In essence, the coupling of the prompt-hippocampus for update and generalization, can work cooperatively to construct the mappings between spatial-temporal contexts and the prediction residuals in traditional learning process (i.e., residuals at warm-up stage).


	\section{Related work}
	\textbf{Spatiotemporal learning.} Great efforts have been made to empower diverse exciting spatiotemporal applications from traffic prediction~\cite{guo2019attention,wang2023pattern,wang2020multi,ouyang2023citytrans,zheng2014urban,liang2021revisiting}, environmental modeling~\cite{liang2023airformer,du2023deciphering,amato2020novel}, to housing price prediction~\cite{liu2013spatial, wang2021joint}. Among them, various grid convolution~\cite{zhang2017deep,ye2019co} or graph convolution networks~\cite{zhou2020foresee, wu2020connecting, yu2018spatio} are devised to capture spatial correlations while temporal convolution~\cite{wu2020connecting} or variants of RNN~\cite{bai2020adaptive,huang2023crossgnn,wang2022predrnn,yao2023modernn,wang2022predrnn,wang2018predrnn++} are well-designed to explicitly model temporal dependencies. Actually, the urban elements and city structures are never static, but the almost all of existing spatiotemporal models assume the same data distribution between training and testing sets. Therefore, it poses great challenges to maintain same performances with the expansion of cities as well as the increases of vehicles.

	\textbf{OOD generalization on spatiotemporal learning.} There are two research lines to counteract the OOD challenges within spatiotemporal forecasting, continuous learning based and the causal perspective based~\cite{yuan2024generative,parisi2018lifelong,chen2021trafficstream,wang2023pattern,zhou2023maintain,xia2023deciphering,yuan2023environment,wang2023brave}. Continuous spatiotemporal learning is devised to update model with new data instance with experience reply~\cite{parisi2018lifelong,chen2021trafficstream,wang2023pattern} where they re-train the partial neural architecture by re-arranging the training set. Specifically, Wang, et, al. proposes a historical-data replay strategy, TrafficStream, to update the neural network with all nodes feeding~\cite{chen2021trafficstream}, while PECPM dynamically manages a spatiotemporal pattern bank with conflict nodes, which reduces the memory storage burdens~\cite{wang2023pattern}. Unfortunately, memory-based methods will inevitably increase the storage space when the network expansion and continuous occurrences of new patterns. To this end, causal-based learning explicitly models temporal environments and captures invariant representations and stable correlations across environments by observing model behaviors. CauSTG reflects complex spatial-temporal dependencies via learnable parameters and transfer the relatively stable weights from historical environment to unseen scenarios~\cite{zhou2023maintain}. CaST~\cite{xia2023deciphering} and EAGLE~\cite{yuan2023environment} explicitly model the environment by disentangling the environment-aware representation and imitate the OOD scenarios via generating new environments. Even prosperity, CauSTG requires multiple training of same neural architectures at different environments, while CaST, EAGLE, and sUrban mimic the perturbation on features and extends the boundary of well-learned data with environment reconstruction-sampling strategy. However, the unseen environments are infinite and the boundary of existing sample space cannot be unlimitedly extended, thus these solutions are still trained a closed space and short of adapting model to new data and environments. To summarize, existing methods for spatiotemporal OOD generalization either requires high computation power, or cannot well deal with new distribution instances, thus lacking capacity in model evolution and adaptation.
	
	\textbf{Learning from neuroscience.} Neuroscience is a discipline investigating how human brain works for remembering, learning, and consolidation~\cite{lindsay2020attention,mathis2020deep,bessadok2022graph,gupta2013artificial,zenke2017continual}.  Early researches have revealed the similarity between machine learning and human studying behavior, where the neural network is initially developed by imitating the brain neural architectures~\cite{zou2009overview} and it will be activated when the information flow exceeds a fixed threshold~\cite{gupta2013artificial}. Biological neural networks can be capable of flexibly modulating synaptic plastic to respond to dynamic inputs. The inspired strategies can be summarized as weight regularization~\cite{zenke2017continual} from stabilization of previously-learned synaptic changes, memory extension and formation from the number expansion and pruning of functional connections~\cite{kirkpatrick2017overcoming, wang2023pattern, van2020brain}, as well as meta-learning, stemmed from activity-dependent synaptic plasticity~\cite{pan2020spatio}. However, these methods cannot well interpret what exact to remember and forget along the learning process. As for a collaborative strategy, the complementary learning system (CLS) theory uncovers that regions in brain consist of complementary functions for remembering knowledge, the hippocampus space to learn new skills fast while neocortex structure to progressively learn stable and long-term knowledge ~\cite{kumaran2016learning}. Various machine learning methods are developed from CLS~\cite{arani2022learning}, to enable a more generalizable model and research evidence show that it shares similarity with continuous learning~\cite{wang2023incorporating,wang2023comprehensive} to adapt multi-task learning. To this end, the structure of the biological brain is sophisticated and informative, exploring the potential structure-reaction of biological brains are valuable to improve machine learning efficiency and potentially address the challenges unsolved currently.
	
	\textbf{Our work.} Rather than taking  efforts on series-level pattern computation~\cite{chen2021trafficstream} and pattern expansion~\cite{wang2023pattern},  or  repeated  training on divided  datasets~\cite{zhou2023maintain}, we couple the complementary learning in neuroscience theory, with spatiotemporal neural architecture to accommodate the streaming and dynamic series observations. This design simultaneously preserves the transferrable stable knowledge from existing data and quickly adapts our model to new arrival instances with new patterns via respectively  updating neocortex and hippocampus neural structures. As a result, our ComS2T can efficiently and effectively empower the spatiotemporal model (even any ST learner) the capacity to evolve with  data distribution shifts countering both structural and temporal shifts in a unified framework.

	\section{Conclusion and discussion}
	In this work, motivated by recent advance in neuroscience, we couple the idea of complementary learning with spatiotemporal forecasting, and propose a prompt-based complementary learning system, ComS2T, to equip the model with data adaptation and evolution capacity. We first decouple the spatial-temporal learning neural network into two disjoint architectures, stable neocortex and dynamic hippocampus, by explicitly modeling the training behavior of learnable weights. To enable efficient model evolution with accessible data, we instantiate additional environments with spatial-temporal prompts to characterize the data distribution over observations and enable prompts learnable with self-supervision signals. Then we can progressively disentangle the neural architecture and incorporate informative prompts into dynamic hippocampus for environment-aware fine-tuning. Our ComS2T especially allows model adaptation conditioned on environment prompts during training stage, thus the fine-tune of prompts can be extended to testing stages when environments correspondingly change, which empowers the model evolution upon new data pattern arrives. Extensive experiments have been conducted on four real-world urban datasets with spatial and temporal shifts. Empirical results demonstrate our ComS2T can counteract the out-of-distribution challenges over urban streaming data, improving performances 0.73\%$\sim$20.70\% and 0.36\%$\sim$17.30\% respectively under temporal and structural shifts. Also, the substantial visualized case studies illustrate the informative and semantic intermediate results and effective disentanglement learning scheme, enhancing the intrinsic interpretability and better understanding of our ComS2T.
	
	\textbf{Further discussion of coupling neuroscience and computer science. } Neuroscience is a discipline full of mystery but of great values. Our ComS2T can be an initial practice of incorporating neuroscience into machine learning system, which takes advantage of the learning schemes in both human brain and artificial neural network. We first inherit the disentangled role in human brain to decouple two disjoint and complementary parameter spaces from spatiotemporal neural architecture, neocortex and hippocampus structures, enabling the consolidation of historical knowledge and quick update of new information. To better imitate the memory system in human brain, we especially borrow the self-supervision from machine learning and design a self-supervised environment prompt, enabling prompts to reflect the update of newly arrived data patterns and guide the update of hippocampus architecture in main learning framework. With this design, our main framework can be sensitive to changes over prompts and the stable neocortex and fine-tuned parameters in hippocampus can be well transferred, resulting in a more efficient and effective data-adaptive learning spatiotemporal learning. The success of this coupling scheme provides insights into developing more generalizable machine learning systems by investigating interesting and practical mechanisms in neuroscience, e.g., how human learn new skills based on similar historical knowledge, and how different neurons in brain work cooperatively for a complex task. We believe these inherent mechanisms can benefit better designs of both artificial neural architecture and training strategies.
	
	\textbf{Future works }can be divided on two-fold. First, we will make consistent efforts on improving the spatiotemporal complementary learning by further alleviating the training-testing efficiency and data alignment issue between environment and main observations. Second, we are going to devote efforts into finding the interesting and practical mechanism in human and animal brains via both experiments and literature, then exploit the phenomenon as well as inherent mechanism as the insights of model designs, leading to a more flexible and intelligent machine learning model countering challenges faced in computer science.

	%

	\ifCLASSOPTIONcompsoc
	\section*{Acknowledgments}
	\else
	\section*{Acknowledgment}
	\fi	
	This paper is partially supported by the National Natural Science Foundation of China (No.62072427, No.12227901), the Project of Stable Support  for Youth Team in Basic Research Field, CAS (No.YSBR-005), the grant from State Key Laboratory of Resources and Environmental Information System,  Guangzhou Municipal Science and Technology Project (2023A03J0011), and  Guangzhou-HKUST (GZ) Joint Funding Program (No. 2024A03J0620).

	\ifCLASSOPTIONcaptionsoff
	\newpage
	\fi

	
	
	%
	\bibliographystyle{IEEEtran}
	\bibliography{caustg}

\begin{thebibliography}{10}
\providecommand{\url}[1]{#1}
\csname url@samestyle\endcsname
\providecommand{\newblock}{\relax}
\providecommand{\bibinfo}[2]{#2}
\providecommand{\BIBentrySTDinterwordspacing}{\spaceskip=0pt\relax}
\providecommand{\BIBentryALTinterwordstretchfactor}{4}
\providecommand{\BIBentryALTinterwordspacing}{\spaceskip=\fontdimen2\font plus
\BIBentryALTinterwordstretchfactor\fontdimen3\font minus
  \fontdimen4\font\relax}
\providecommand{\BIBforeignlanguage}[2]{{%
\expandafter\ifx\csname l@#1\endcsname\relax
\typeout{** WARNING: IEEEtran.bst: No hyphenation pattern has been}%
\typeout{** loaded for the language `#1'. Using the pattern for}%
\typeout{** the default language instead.}%
\else
\language=\csname l@#1\endcsname
\fi
#2}}
\providecommand{\BIBdecl}{\relax}
\BIBdecl

\bibitem{wu2020connecting}
Z.~Wu, S.~Pan, G.~Long, J.~Jiang, X.~Chang, and C.~Zhang, ``Connecting the
  dots: Multivariate time series forecasting with graph neural networks,'' in
  \emph{KDD}, 2020, pp. 753--763.

\bibitem{huang2023crossgnn}
Q.~Huang, L.~Shen, R.~Zhang, S.~Ding, B.~Wang, Z.~Zhou, and Y.~Wang,
  ``Crossgnn: Confronting noisy multivariate time series via cross interaction
  refinement,'' in \emph{Thirty-seventh Conference on Neural Information
  Processing Systems}, 2023.

\bibitem{zheng2023diffuflow}
Y.~Zheng, L.~Zhong, S.~Wang, Y.~Yang, W.~Gu, J.~Zhang, and J.~Wang,
  ``Diffuflow: Robust fine-grained urban flow inference with denoising
  diffusion model,'' in \emph{Proceedings of the 32nd ACM International
  Conference on Information and Knowledge Management}, 2023, pp. 3505--3513.

\bibitem{jin2023spatio}
G.~Jin, Y.~Liang, Y.~Fang, J.~Huang, J.~Zhang, and Y.~Zheng, ``Spatio-temporal
  graph neural networks for predictive learning in urban computing: A survey,''
  \emph{arXiv preprint arXiv:2303.14483}, 2023.

\bibitem{wang2020deep}
S.~Wang, J.~Cao, and S.~Y. Philip, ``Deep learning for spatio-temporal data
  mining: A survey,'' \emph{IEEE transactions on knowledge and data
  engineering}, vol.~34, no.~8, pp. 3681--3700, 2020.

\bibitem{lu2022spatio}
B.~Lu, X.~Gan, W.~Zhang, H.~Yao, L.~Fu, and X.~Wang, ``Spatio-temporal graph
  few-shot learning with cross-city knowledge transfer,'' in \emph{Proceedings
  of the 28th ACM SIGKDD Conference on Knowledge Discovery and Data Mining},
  2022, pp. 1162--1172.

\bibitem{pan2020spatio}
Z.~Pan, W.~Zhang, Y.~Liang, W.~Zhang, Y.~Yu, J.~Zhang, and Y.~Zheng,
  ``Spatio-temporal meta learning for urban traffic prediction,'' \emph{IEEE
  Transactions on Knowledge and Data Engineering}, vol.~34, no.~3, pp.
  1462--1476, 2020.

\bibitem{liu2023itransformer}
Y.~Liu, T.~Hu, H.~Zhang, H.~Wu, S.~Wang, L.~Ma, and M.~Long, ``itransformer:
  Inverted transformers are effective for time series forecasting,'' in
  \emph{The Twelfth International Conference on Learning Representations},
  2023.

\bibitem{dong2024simmtm}
J.~Dong, H.~Wu, H.~Zhang, L.~Zhang, J.~Wang, and M.~Long, ``Simmtm: A simple
  pre-training framework for masked time-series modeling,'' \emph{Advances in
  Neural Information Processing Systems}, vol.~36, 2024.

\bibitem{wu2022timesnet}
H.~Wu, T.~Hu, Y.~Liu, H.~Zhou, J.~Wang, and M.~Long, ``Timesnet: Temporal
  2d-variation modeling for general time series analysis,'' in \emph{The
  Eleventh International Conference on Learning Representations}, 2022.

\bibitem{liu2022non}
Y.~Liu, H.~Wu, J.~Wang, and M.~Long, ``Non-stationary transformers: Rethinking
  the stationarity in time series forecasting,'' \emph{arXiv preprint
  arXiv:2205.14415}, 2022.

\bibitem{zhou2020foresee}
Z.~Zhou, Y.~Wang, X.~Xie, L.~Chen, and C.~Zhu, ``Foresee urban sparse traffic
  accidents: A spatiotemporal multi-granularity perspective,'' \emph{IEEE
  TKDE}, 2020.

\bibitem{ji2023spatio}
J.~Ji, J.~Wang, C.~Huang, J.~Wu, B.~Xu, Z.~Wu, J.~Zhang, and Y.~Zheng,
  ``Spatio-temporal self-supervised learning for traffic flow prediction,'' in
  \emph{Proceedings of the AAAI Conference on Artificial Intelligence}, 2023.

\bibitem{li2018diffusion}
Y.~Li, R.~Yu, C.~Shahabi, and Y.~Liu, ``Diffusion convolutional recurrent
  neural network: Data-driven traffic forecasting,'' in \emph{International
  Conference on Learning Representations}, 2018.

\bibitem{zhou2020riskoracle}
Z.~Zhou, Y.~Wang, X.~Xie, L.~Chen, and H.~Liu, ``Riskoracle: A minute-level
  citywide traffic accident forecasting framework,'' in \emph{Proceedings of
  the AAAI conference on artificial intelligence}, vol.~34, no.~01, 2020, pp.
  1258--1265.

\bibitem{wu2023interpretable}
H.~Wu, H.~Zhou, M.~Long, and J.~Wang, ``Interpretable weather forecasting for
  worldwide stations with a unified deep model,'' \emph{Nature Machine
  Intelligence}, pp. 1--10, 2023.

\bibitem{castro2021stconvs2s}
R.~Castro, Y.~M. Souto, E.~Ogasawara, F.~Porto, and E.~Bezerra, ``Stconvs2s:
  Spatiotemporal convolutional sequence to sequence network for weather
  forecasting,'' \emph{Neurocomputing}, vol. 426, pp. 285--298, 2021.

\bibitem{chen2023fengwu}
K.~Chen, T.~Han, J.~Gong, L.~Bai, F.~Ling, J.-J. Luo, X.~Chen, L.~Ma, T.~Zhang,
  R.~Su \emph{et~al.}, ``Fengwu: Pushing the skillful global medium-range
  weather forecast beyond 10 days lead,'' \emph{arXiv preprint
  arXiv:2304.02948}, 2023.

\bibitem{zhang2023skilful}
Y.~Zhang, M.~Long, K.~Chen, L.~Xing, R.~Jin, M.~I. Jordan, and J.~Wang,
  ``Skilful nowcasting of extreme precipitation with nowcastnet,''
  \emph{Nature}, vol. 619, no. 7970, pp. 526--532, 2023.

\bibitem{liang2023airformer}
Y.~Liang, Y.~Xia, S.~Ke, Y.~Wang, Q.~Wen, J.~Zhang, Y.~Zheng, and
  R.~Zimmermann, ``Airformer: Predicting nationwide air quality in china with
  transformers,'' in \emph{Proceedings of the AAAI Conference on Artificial
  Intelligence}, vol.~37, no.~12, 2023, pp. 14\,329--14\,337.

\bibitem{du2023deciphering}
W.~Du, L.~Chen, H.~Wang, Z.~Shan, Z.~Zhou, W.~Li, and Y.~Wang, ``Deciphering
  urban traffic impacts on air quality by deep learning and emission
  inventory,'' \emph{Journal of Environmental Sciences}, vol. 124, pp.
  745--757, 2023.

\bibitem{du2019deep}
S.~Du, T.~Li, Y.~Yang, and S.-J. Horng, ``Deep air quality forecasting using
  hybrid deep learning framework,'' \emph{IEEE Transactions on Knowledge and
  Data Engineering}, vol.~33, no.~6, pp. 2412--2424, 2019.

\bibitem{chen2023group}
L.~Chen, J.~Xu, B.~Wu, and J.~Huang, ``Group-aware graph neural network for
  nationwide city air quality forecasting,'' \emph{ACM Transactions on
  Knowledge Discovery from Data}, vol.~18, no.~3, pp. 1--20, 2023.

\bibitem{wang2023brave}
K.~Wang, Y.~Liang, X.~Li, G.~Li, B.~Ghanem, R.~Zimmermann, H.~Yi, Y.~Zhang,
  Y.~Wang \emph{et~al.}, ``Brave the wind and the waves: Discovering robust and
  generalizable graph lottery tickets,'' \emph{IEEE Transactions on Pattern
  Analysis and Machine Intelligence}, 2023.

\bibitem{wu2021discovering}
Y.~Wu, X.~Wang, A.~Zhang, X.~He, and T.-S. Chua, ``Discovering invariant
  rationales for graph neural networks,'' in \emph{International Conference on
  Learning Representations}, 2021.

\bibitem{wu2021handling}
Q.~Wu, H.~Zhang, J.~Yan, and D.~Wipf, ``Handling distribution shifts on graphs:
  An invariance perspective,'' in \emph{International Conference on Learning
  Representations}, 2022.

\bibitem{li2022let}
S.~Li, X.~Wang, A.~Zhang, Y.~Wu, X.~He, and T.-S. Chua, ``Let invariant
  rationale discovery inspire graph contrastive learning,'' in
  \emph{International Conference on Machine Learning}.\hskip 1em plus 0.5em
  minus 0.4em\relax PMLR, 2022, pp. 13\,052--13\,065.

\bibitem{du2021adarnn}
Y.~Du, J.~Wang, W.~Feng, S.~Pan, T.~Qin, R.~Xu, and C.~Wang, ``Adarnn: Adaptive
  learning and forecasting of time series,'' in \emph{Proceedings of the 30th
  ACM International Conference on Information \& Knowledge Management}, 2021,
  pp. 402--411.

\bibitem{zhou2023maintain}
Z.~Zhou, Q.~Huang, K.~Yang, K.~Wang, X.~Wang, Y.~Zhang, Y.~Liang, and Y.~Wang,
  ``Maintaining the status quo: Capturing invariant relations for ood
  spatiotemporal learning,'' in \emph{Proceedings of the 29th ACM SIGKDD
  Conference on Knowledge Discovery and Data Mining}, 2023, p. 3603–3614.

\bibitem{chen2021trafficstream}
X.~Chen, J.~Wang, and K.~Xie, ``Trafficstream: A streaming traffic flow
  forecasting framework based on graph neural networks and continual
  learning,'' \emph{arXiv preprint arXiv:2106.06273}, 2021.

\bibitem{wang2023pattern}
B.~Wang, Y.~Zhang, X.~Wang, P.~Wang, Z.~Zhou, L.~Bai, and Y.~Wang, ``Pattern
  expansion and consolidation on evolving graphs for continual traffic
  prediction,'' in \emph{Proceedings of the 29th ACM SIGKDD Conference on
  Knowledge Discovery and Data Mining}, 2023, pp. 2223--2232.

\bibitem{wang2023surban}
Q.~Wang, B.~Guo, L.~Cheng, and Z.~Yu, ``surban: Stable prediction for unseen
  urban data from location-based sensors,'' \emph{Proceedings of the ACM on
  Interactive, Mobile, Wearable and Ubiquitous Technologies}, vol.~7, no.~3,
  pp. 1--20, 2023.

\bibitem{xia2023deciphering}
Y.~Xia, Y.~Liang, H.~Wen, X.~Liu, K.~Wang, Z.~Zhou, and R.~Zimmermann,
  ``Deciphering spatio-temporal graph forecasting: A causal lens and
  treatment,'' in \emph{Thirty-seventh Conference on Neural Information
  Processing Systems}, 2023.

\bibitem{yuan2023environment}
H.~Yuan, Q.~Sun, X.~Fu, Z.~Zhang, C.~Ji, H.~Peng, and J.~Li,
  ``Environment-aware dynamic graph learning for out-of-distribution
  generalization,'' in \emph{Thirty-seventh Conference on Neural Information
  Processing Systems}, 2023.

\bibitem{o2014complementary}
R.~C. O’Reilly, R.~Bhattacharyya, M.~D. Howard, and N.~Ketz, ``Complementary
  learning systems,'' \emph{Cognitive science}, vol.~38, no.~6, pp. 1229--1248,
  2014.

\bibitem{kumaran2016learning}
D.~Kumaran, D.~Hassabis, and J.~L. McClelland, ``What learning systems do
  intelligent agents need? complementary learning systems theory updated,''
  \emph{Trends in cognitive sciences}, vol.~20, no.~7, pp. 512--534, 2016.

\bibitem{mcclelland2020integration}
J.~L. McClelland, B.~L. McNaughton, and A.~K. Lampinen, ``Integration of new
  information in memory: new insights from a complementary learning systems
  perspective,'' \emph{Philosophical Transactions of the Royal Society B}, vol.
  375, no. 1799, p. 20190637, 2020.

\bibitem{lee2020clinical}
C.~S. Lee and A.~Y. Lee, ``Clinical applications of continual learning machine
  learning,'' \emph{The Lancet Digital Health}, vol.~2, no.~6, pp. e279--e281,
  2020.

\bibitem{arani2022learning}
E.~Arani, F.~Sarfraz, and B.~Zonooz, ``Learning fast, learning slow: A general
  continual learning method based on complementary learning system,'' in
  \emph{International Conference on Learning Representations}, 2022.

\bibitem{zhang2023promptst}
Z.~Zhang, X.~Zhao, Q.~Liu, C.~Zhang, Q.~Ma, W.~Wang, H.~Zhao, Y.~Wang, and
  Z.~Liu, ``Promptst: Prompt-enhanced spatio-temporal multi-attribute
  prediction,'' in \emph{Proceedings of the 32nd ACM International Conference
  on Information and Knowledge Management}, 2023, pp. 3195--3205.

\bibitem{liu2022p}
X.~Liu, K.~Ji, Y.~Fu, W.~Tam, Z.~Du, Z.~Yang, and J.~Tang, ``P-tuning: Prompt
  tuning can be comparable to fine-tuning across scales and tasks,'' in
  \emph{Proceedings of the 60th Annual Meeting of the Association for
  Computational Linguistics (Volume 2: Short Papers)}, 2022, pp. 61--68.

\bibitem{guo2017deepfm}
H.~Guo, T.~Ruiming, Y.~Ye, Z.~Li, and X.~He, ``Deepfm: A factorization-machine
  based neural network for ctr prediction,'' in \emph{Proceedings of the
  Twenty-Sixth International Joint Conference on Artificial
  Intelligence}.\hskip 1em plus 0.5em minus 0.4em\relax International Joint
  Conferences on Artificial Intelligence Organization, 2017.

\bibitem{wang2020pm2}
S.~Wang, Y.~Li, J.~Zhang, Q.~Meng, L.~Meng, and F.~Gao, ``Pm2. 5-gnn: A domain
  knowledge enhanced graph neural network for pm2. 5 forecasting,'' in
  \emph{Proceedings of the 28th International Conference on Advances in
  Geographic Information Systems}, 2020, pp. 163--166.

\bibitem{wu2019graph}
Z.~Wu, S.~Pan, G.~Long, J.~Jiang, and C.~Zhang, ``Graph wavenet for deep
  spatial-temporal graph modeling,'' in \emph{Proceedings of the 28th
  International Joint Conference on Artificial Intelligence}, 2019, pp.
  1907--1913.

\bibitem{guo2019attention}
S.~Guo, Y.~Lin, N.~Feng, C.~Song, and H.~Wan, ``Attention based
  spatial-temporal graph convolutional networks for traffic flow forecasting,''
  in \emph{AAAI}, vol.~33, no.~01, 2019, pp. 922--929.

\bibitem{wang2020multi}
S.~Wang, H.~Miao, H.~Chen, and Z.~Huang, ``Multi-task adversarial
  spatial-temporal networks for crowd flow prediction,'' in \emph{Proceedings
  of the 29th ACM international conference on information \& knowledge
  management}, 2020, pp. 1555--1564.

\bibitem{ouyang2023citytrans}
X.~Ouyang, Y.~Yang, W.~Zhou, Y.~Zhang, H.~Wang, and W.~Huang, ``Citytrans:
  Domain-adversarial training with knowledge transfer for spatio-temporal
  prediction across cities,'' \emph{IEEE Transactions on Knowledge and Data
  Engineering}, 2023.

\bibitem{zheng2014urban}
Y.~Zheng, L.~Capra, O.~Wolfson, and H.~Yang, ``Urban computing: concepts,
  methodologies, and applications,'' \emph{ACM Transactions on Intelligent
  Systems and Technology (TIST)}, vol.~5, no.~3, pp. 1--55, 2014.

\bibitem{liang2021revisiting}
Y.~Liang, K.~Ouyang, Y.~Wang, Y.~Liu, J.~Zhang, Y.~Zheng, and D.~S. Rosenblum,
  ``Revisiting convolutional neural networks for citywide crowd flow
  analytics,'' in \emph{Machine Learning and Knowledge Discovery in Databases:
  European Conference, ECML PKDD 2020, Ghent, Belgium, September 14--18, 2020,
  Proceedings, Part I}.\hskip 1em plus 0.5em minus 0.4em\relax Springer, 2021,
  pp. 578--594.

\bibitem{amato2020novel}
F.~Amato, F.~Guignard, S.~Robert, and M.~Kanevski, ``A novel framework for
  spatio-temporal prediction of environmental data using deep learning,''
  \emph{Scientific reports}, vol.~10, no.~1, p. 22243, 2020.

\bibitem{liu2013spatial}
X.~Liu, ``Spatial and temporal dependence in house price prediction,''
  \emph{The Journal of Real Estate Finance and Economics}, vol.~47, no.~2, pp.
  341--369, 2013.

\bibitem{wang2021joint}
P.~Wang, C.~Ge, Z.~Zhou, X.~Wang, Y.~Li, and Y.~Wang, ``Joint gated
  co-attention based multi-modal networks for subregion house price
  prediction,'' \emph{IEEE Transactions on Knowledge and Data Engineering},
  2021.

\bibitem{zhang2017deep}
J.~Zhang, Y.~Zheng, and D.~Qi, ``Deep spatio-temporal residual networks for
  citywide crowd flows prediction,'' in \emph{Thirty-first AAAI conference on
  artificial intelligence}, 2017.

\bibitem{ye2019co}
J.~Ye, L.~Sun, B.~Du, Y.~Fu, X.~Tong, and H.~Xiong, ``Co-prediction of multiple
  transportation demands based on deep spatio-temporal neural network,'' in
  \emph{Proceedings of the 25th ACM SIGKDD international conference on
  knowledge discovery \& data mining}, 2019, pp. 305--313.

\bibitem{yu2018spatio}
B.~Yu, H.~Yin, and Z.~Zhu, ``Spatio-temporal graph convolutional networks: a
  deep learning framework for traffic forecasting,'' in \emph{IJCAI}, 2018, pp.
  3634--3640.

\bibitem{bai2020adaptive}
L.~Bai, L.~Yao, C.~Li, X.~Wang, and C.~Wang, ``Adaptive graph convolutional
  recurrent network for traffic forecasting,'' \emph{NIPS}, vol.~33, 2020.

\bibitem{wang2022predrnn}
Y.~Wang, H.~Wu, J.~Zhang, Z.~Gao, J.~Wang, S.~Y. Philip, and M.~Long,
  ``Predrnn: A recurrent neural network for spatiotemporal predictive
  learning,'' \emph{IEEE Transactions on Pattern Analysis and Machine
  Intelligence}, vol.~45, no.~2, pp. 2208--2225, 2022.

\bibitem{yao2023modernn}
Z.~Yao, Y.~Wang, H.~Wu, J.~Wang, and M.~Long, ``Modernn: Harnessing
  spatiotemporal mode collapse in unsupervised predictive learning,''
  \emph{IEEE Transactions on Pattern Analysis and Machine Intelligence}, 2023.

\bibitem{wang2018predrnn++}
Y.~Wang, Z.~Gao, M.~Long, J.~Wang, and S.~Y. Philip, ``Predrnn++: Towards a
  resolution of the deep-in-time dilemma in spatiotemporal predictive
  learning,'' in \emph{International Conference on Machine Learning}.\hskip 1em
  plus 0.5em minus 0.4em\relax PMLR, 2018, pp. 5123--5132.

\bibitem{yuan2024generative}
Y.~Yuan, C.~Shao, J.~Ding, D.~Jin, and Y.~Li, ``A generative pre-training
  framework for spatio-temporal graph transfer learning,'' \emph{arXiv preprint
  arXiv:2402.11922}, 2024.

\bibitem{parisi2018lifelong}
G.~I. Parisi, J.~Tani, C.~Weber, and S.~Wermter, ``Lifelong learning of
  spatiotemporal representations with dual-memory recurrent
  self-organization,'' \emph{Frontiers in neurorobotics}, vol.~12, p.~78, 2018.

\bibitem{lindsay2020attention}
G.~W. Lindsay, ``Attention in psychology, neuroscience, and machine learning,''
  \emph{Frontiers in computational neuroscience}, vol.~14, p.~29, 2020.

\bibitem{mathis2020deep}
M.~W. Mathis and A.~Mathis, ``Deep learning tools for the measurement of animal
  behavior in neuroscience,'' \emph{Current opinion in neurobiology}, vol.~60,
  pp. 1--11, 2020.

\bibitem{bessadok2022graph}
A.~Bessadok, M.~A. Mahjoub, and I.~Rekik, ``Graph neural networks in network
  neuroscience,'' \emph{IEEE Transactions on Pattern Analysis and Machine
  Intelligence}, vol.~45, no.~5, pp. 5833--5848, 2022.

\bibitem{gupta2013artificial}
N.~Gupta \emph{et~al.}, ``Artificial neural network,'' \emph{Network and
  Complex Systems}, vol.~3, no.~1, pp. 24--28, 2013.

\bibitem{zenke2017continual}
F.~Zenke, B.~Poole, and S.~Ganguli, ``Continual learning through synaptic
  intelligence,'' in \emph{International conference on machine learning}.\hskip
  1em plus 0.5em minus 0.4em\relax PMLR, 2017, pp. 3987--3995.

\bibitem{zou2009overview}
J.~Zou, Y.~Han, and S.-S. So, ``Overview of artificial neural networks,''
  \emph{Artificial neural networks: methods and applications}, pp. 14--22,
  2009.

\bibitem{kirkpatrick2017overcoming}
J.~Kirkpatrick, R.~Pascanu, N.~Rabinowitz, J.~Veness, G.~Desjardins, A.~A.
  Rusu, K.~Milan, J.~Quan, T.~Ramalho, A.~Grabska-Barwinska \emph{et~al.},
  ``Overcoming catastrophic forgetting in neural networks,'' \emph{Proceedings
  of the national academy of sciences}, vol. 114, no.~13, pp. 3521--3526, 2017.

\bibitem{van2020brain}
G.~M. Van~de Ven, H.~T. Siegelmann, and A.~S. Tolias, ``Brain-inspired replay
  for continual learning with artificial neural networks,'' \emph{Nature
  communications}, vol.~11, no.~1, p. 4069, 2020.

\bibitem{wang2023incorporating}
L.~Wang, X.~Zhang, Q.~Li, M.~Zhang, H.~Su, J.~Zhu, and Y.~Zhong,
  ``Incorporating neuro-inspired adaptability for continual learning in
  artificial intelligence,'' \emph{Nature Machine Intelligence}, pp. 1--13,
  2023.

\bibitem{wang2023comprehensive}
L.~Wang, X.~Zhang, H.~Su, and J.~Zhu, ``A comprehensive survey of continual
  learning: Theory, method and application,'' \emph{arXiv preprint
  arXiv:2302.00487}, 2023.

\end{thebibliography}

	\section{Biography Section}
	\vspace{-0.5in}
	\begin{IEEEbiography}[{\includegraphics[width=1in,height=1.25in,clip,keepaspectratio]{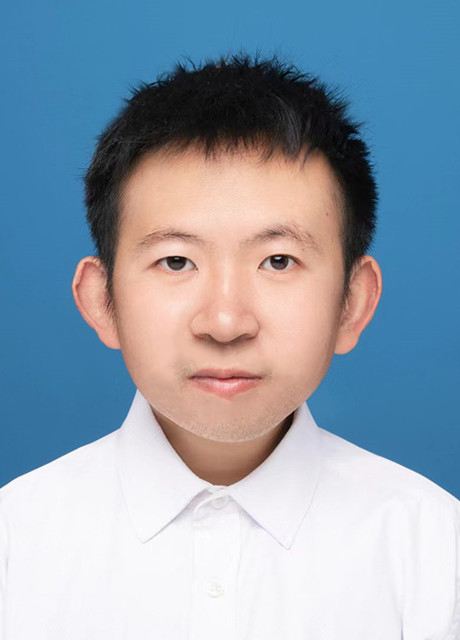}}]{Zhengyang Zhou} is now an associate researcher at Suzhou Institute for Advanced Research, University of Science and Technology of China (USTC). He got his Ph.D. degree at University of Science and Technology of China in 2023. He has published over 30 papers on top conferences and journals such as NeurIPS, ICLR,  KDD, TKDE, WWW, AAAI and ICDE. His mainly research interests include spatiotemporal data minining,   human-centered urban computing, and deep learning  generalization with model behavior analysis. He is now especially  interested in improving  generalization capacity of neural networks for  streaming and  spatiotemporal data.    
		
	\end{IEEEbiography}
	\vspace{-0.25in}
	\begin{IEEEbiography}[{\includegraphics[width=1in,height=1.25in,clip,keepaspectratio]{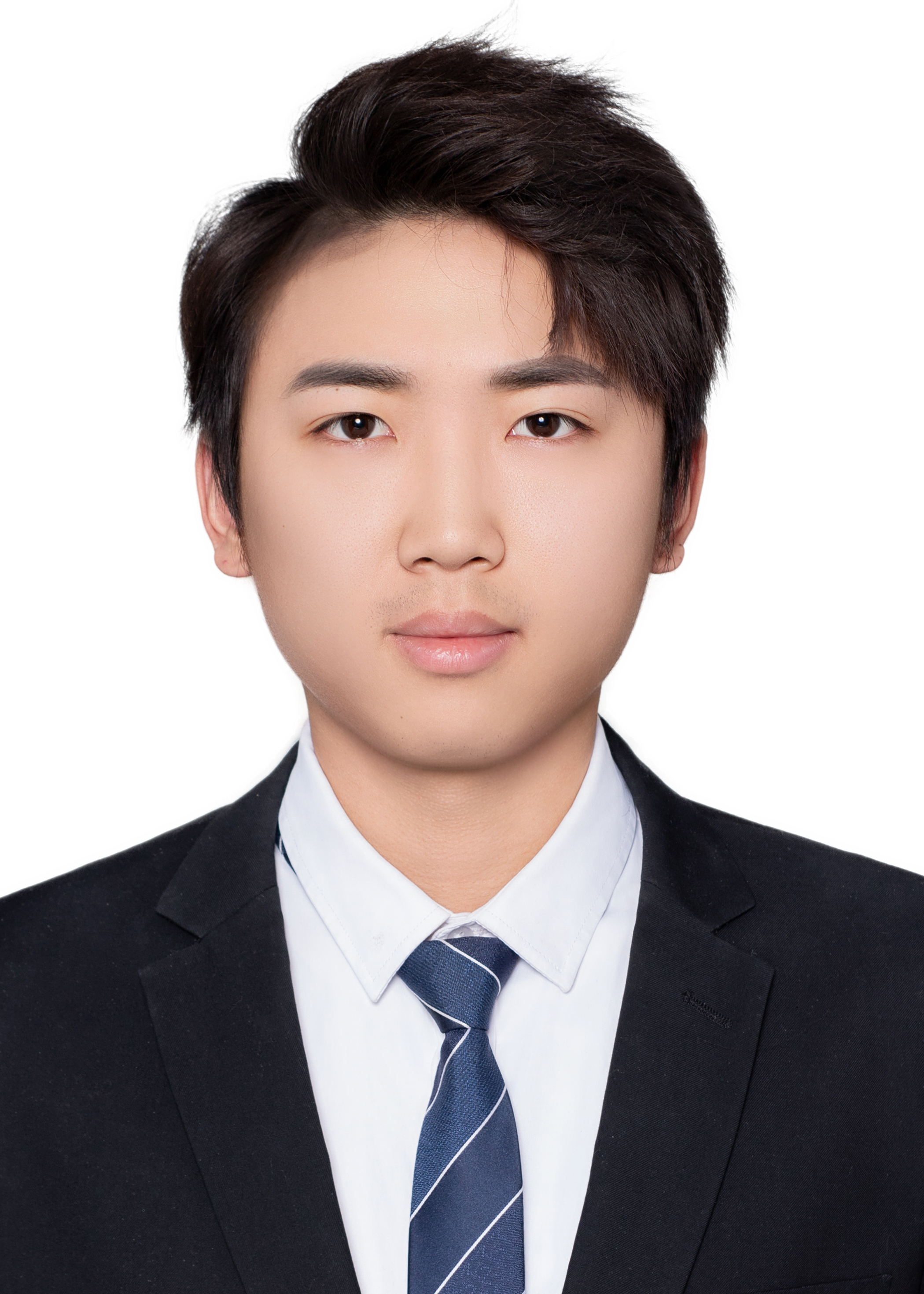}}]{Qihe Huang} is currently pursuing a M.S. degree at University of Science and Technology of China. He obtained his B.S. degree from Nanjing University of Information Science and Technology in 2022. His research centers on spatiotemporal data mining and time-series  forecasting.
	\end{IEEEbiography}
	\vspace{-0.25in}
	\begin{IEEEbiography}[{\includegraphics[width=1in,height=1.25in,clip,keepaspectratio]{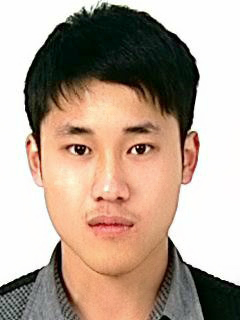}}]{Binwu Wang} is now a PhD candidate at School of Data Science, USTC. His research interests include spatial-temporal data mining and human-centered urban computing. He has published over 10 refereed journal and conference papers in the field of data mining, including AAAI, ICLR, IEEE TITS, DASFAA, IEEE ICDM, WSDM, etc.  
	\end{IEEEbiography}
	\vspace{-0.25in}
	\begin{IEEEbiography}[{\includegraphics[width=1in,height=1.25in,clip,keepaspectratio]{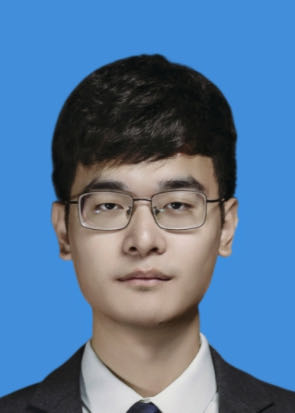}}]{Jianpeng Hou} is currently pursuing a M.S. degree at University of Science and Technology of China. He obtained his B.S. degree from Shandong University in 2023. His research centers on  OOD generalization.
	\end{IEEEbiography}
	\vspace{-0.25in}
	\begin{IEEEbiography}[{\includegraphics[width=1in,height=1.25in,clip,keepaspectratio]{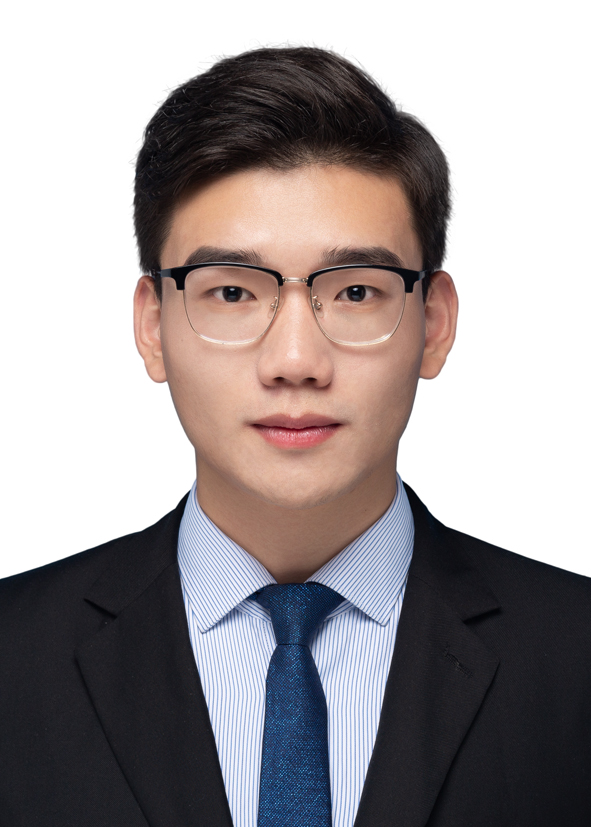}}]{Kuo Yang} received his BE degree from Northeastern University at Qinhuangdao, China, in 2021. He is now working toward  a  PhD degree at School of Data Science, USTC. His mainly research interests are graph representation learning, spatiotemporal data mining and especially subgraph-driven spatiotemporal graph learning.
	\end{IEEEbiography}
	\vspace{-0.25in}
	\begin{IEEEbiography}[{\includegraphics[width=1in,height=1.25in,clip,keepaspectratio]{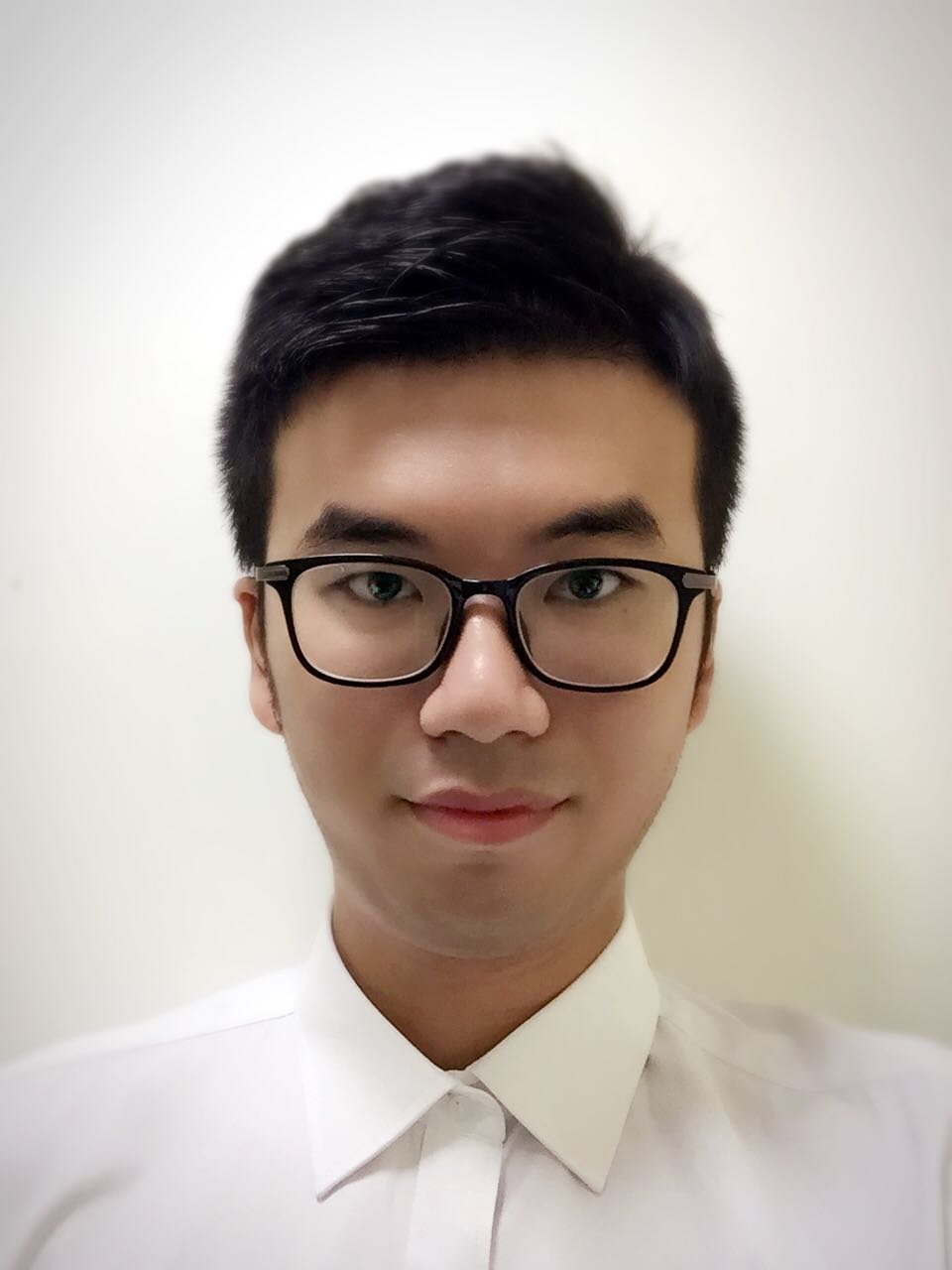}}]{Yuxuan Liang} is an Assistant Professor at Intelligent Transportation Thrust, Hong Kong University of Science and Technology (Guangzhou). He is currently working on the research, development, and innovation of spatio-temporal data mining and AI, with a broad range of applications in smart cities. Prior to that, he obtained his PhD degree at NUS. He published over 40 peer-reviewed papers in refereed journals and conferences, such as KDD, WWW, NeurIPS, ICLR, ECCV, AAAI, IJCAI, Ubicomp, and TKDE. Those papers have been cited over 2,100 times (Google Scholar H-Index: 21). He was recognized as 1 out of 10 most innovative and impactful PhD students focusing on data science in Singapore by Singapore Data Science Consortium (SDSC).
	\end{IEEEbiography}
	\vspace{-0.5in}
	\begin{IEEEbiography}[{\includegraphics[width=1in,height=1.25in,clip,keepaspectratio]{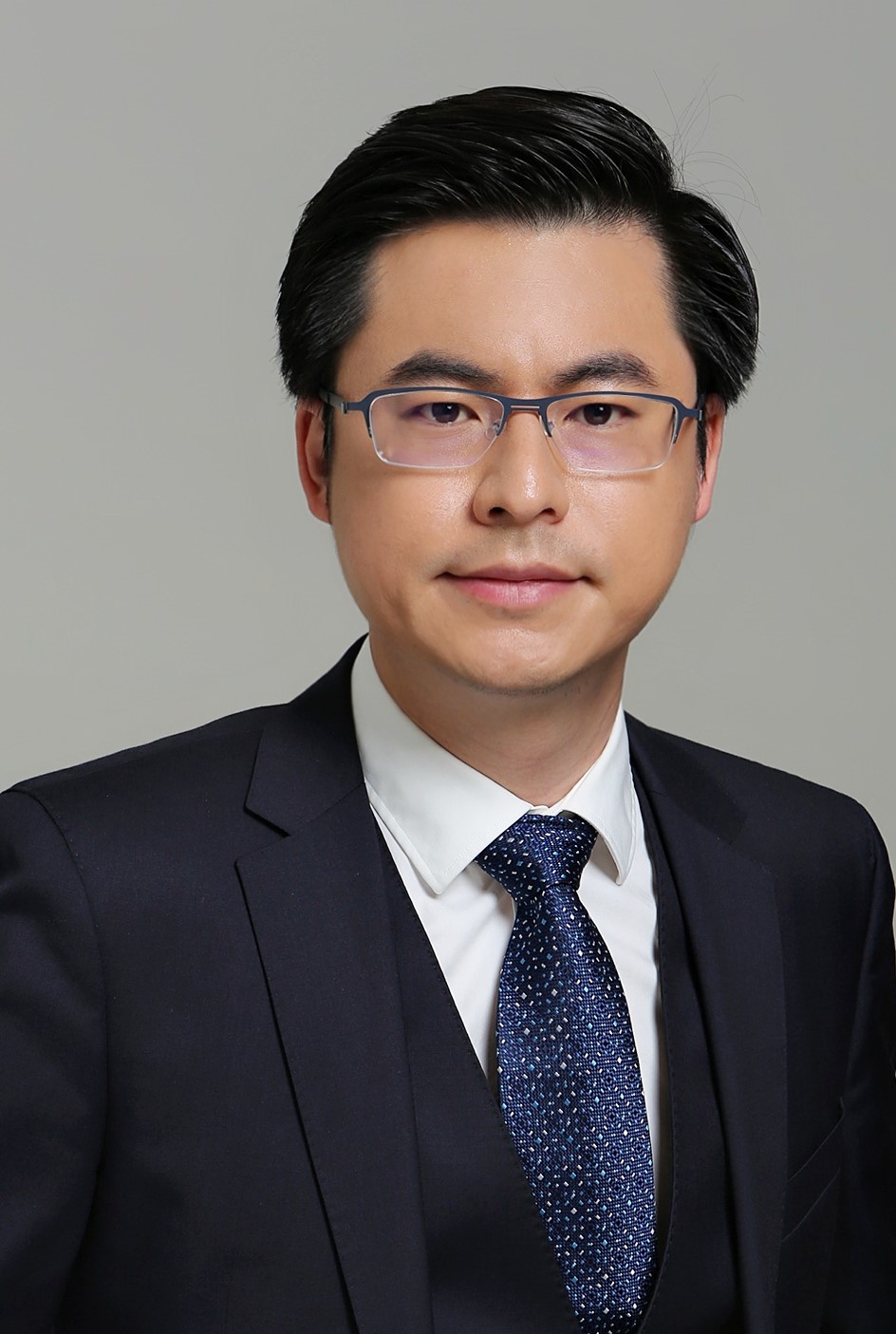}}]{Yu Zheng} is the Vice President of JD.COM and head JD Intelligent Cities Research. Before Joining JD.COM, he was a senior research manager at Microsoft Research. He currently serves as the Editor-in-Chief of ACM Transactions on Intelligent Systems and Technology and has served as the program co-chair of ICDE 2014 (Industrial Track), CIKM 2017 (Industrial Track) and IJCAI 2019 (industrial track). He is also a keynote speaker of AAAI 2019, KDD 2019 Plenary Keynote Panel and IJCAI 2019 Industrial Days. His monograph, entitled Urban Computing, has been used as the first text book in this field. In 2013, he was named one of the Top Innovators under 35 by MIT Technology Review (TR35) and featured by Time Magazine for his research on urban computing. In 2016, Zheng was named an ACM Distinguished Scientist and elevated to an IEEE Fellow in 2020 for his contributions to spatio-temporal data mining and urban computing.
	\end{IEEEbiography}
	\vspace{-0.5in}
	\begin{IEEEbiography}[{\includegraphics[width=1in,height=1.25in,clip,keepaspectratio]{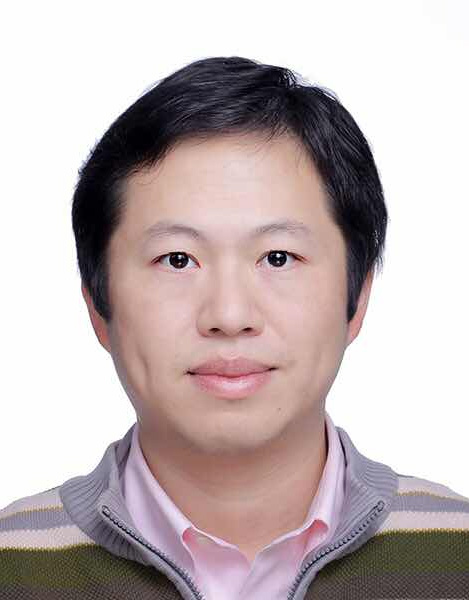}}]{Yang Wang} is now an associate professor at School of Computer Science and Technology, School of Software Engineering, and School of Data Science in USTC. He got his Ph.D. degree at University of Science and Technology of China in 2007. Since then, he keeps working at USTC till now as a postdoc and an associate professor successively. Meanwhile, he also serves as the vice dean of school of software engineering of USTC. His research interest mainly includes wireless (sensor) networks, distribute systems, data mining, and machine learning, and he is also interested in all kinds of applications of AI and data mining technologies especially in urban computing and AI4Science.
	\end{IEEEbiography}
	\vspace{-0.5in}
	\vspace{11pt}
	

	\vspace{11pt}

	\vfill
	
\end{document}